%% file: diagnet.tex
\PassOptionsToPackage{table,dvipsnames}{xcolor}
\documentclass[10pt, conference, letterpaper]{IEEEtran}
\usepackage{cite}
\usepackage{amsmath,amssymb,amsfonts}
\usepackage[utf8]{inputenc}
\usepackage[us,nodayofweek]{datetime}
\usepackage{pifont}
\usepackage{graphicx}
\usepackage{booktabs}
\usepackage{hyperref}
\usepackage{listings}
\usepackage{mathtools}
\usepackage{standalone}
\usepackage{csvsimple}
\usepackage{xspace}
\usepackage{tikz}
\usepackage{pgfplots}
\usepackage{pgfplotstable}
\usepackage{colortbl}
\pgfplotsset{compat=1.14}
\usepgfplotslibrary{groupplots}
\usetikzlibrary{shapes,calc,positioning,backgrounds,patterns,fit,decorations.pathreplacing}
\usepackage[linesnumbered,ruled,vlined]{algorithm2e}
\hypersetup{colorlinks,
  linkcolor={red!50!black},
  citecolor={blue!50!black},
  urlcolor={blue!80!black}
}

\DeclareRobustCommand{\landmarkIcon}{%
  \tikz[baseline=-3pt]{\node[draw,circle,inner sep=0.5pt]{\includegraphics[width=2mm]{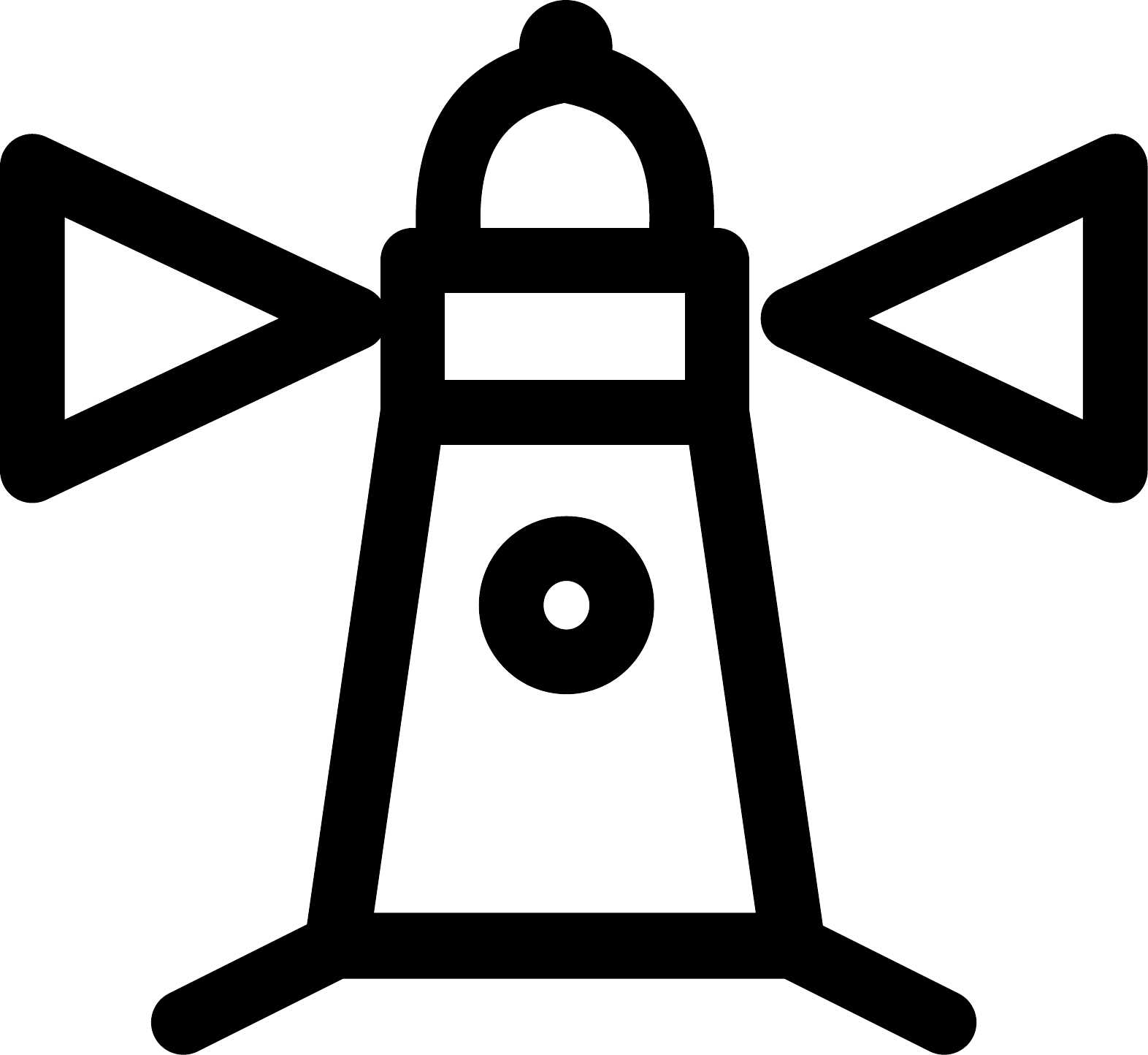}}}
}

\DeclareRobustCommand{\circled}[1]{%
  {\tikz[baseline=(char.base)]{\node[shape=circle,draw,inner sep=0.6pt] (char) {\fontsize{7pt}{7pt}\selectfont #1};}}\hspace{-2pt}
}

\DeclareMathOperator*{\argmax}{arg\,max}

\definecolor{darkgreen}{RGB}{0,120,0}
\definecolor{darkred}{RGB}{200,80,80}
\definecolor{lightg}{RGB}{200,200,200}
\newcommand{\diagnet}{\textsc{Diag\-Net}\xspace}

\makeatletter
\newcommand\resetstackedplots{
\makeatletter
\pgfplots@stacked@isfirstplottrue
\makeatother
\addplot [forget plot,draw=none] coordinates{(0,1)};
}

\begin{document}

\title{\diagnet: towards a generic, Internet-scale \\ root cause analysis solution}

\author{%
  \IEEEauthorblockN{Loïck Bonniot}
  \IEEEauthorblockA{InterDigital \\ Univ Rennes, Inria, CNRS, IRISA \\ loick.bonniot@interdigital.com}
  \and
  \IEEEauthorblockN{Christoph Neumann}
  \IEEEauthorblockA{InterDigital \\ christoph.neumann@interdigital.com}
  \and
  \IEEEauthorblockN{François Taïani}
  \IEEEauthorblockA{Univ Rennes, Inria, CNRS, IRISA \\ francois.taiani@irisa.fr}
}

\maketitle

\begin{abstract}
  \input{abstract}
\end{abstract}

\section{Introduction}

Both content providers and Internet service providers (ISPs) strive to provide the best service to their customers, and allocate very significant resources to diagnose and troubleshoot end-user problems.
For instance, an ISP should ideally be able to immediately detect and explain a service degradation to its users.
Unfortunately, the reason for a problem might lie anywhere between the customer's home and the final data center, and many of the locations involved are not controlled by the ISPs.
Worse, as services grow more complex and interdependent, it is becoming\ increasingly hard to ascertain whether a given perturbation somewhere in the Internet is the cause of a customer's trouble,
causing ISPs' customer support to often struggle to diagnose the root cause of an incident~\cite{Sundaresan2013,Dimopoulos2015}.
Similarly, content providers closely monitor the Quality of Experience (QoE) of their users across the globe, and seek to rapidly resolve any observed degradation, as even a small drop in QoE can have a tremendous impact in revenue and brand image~\cite{Joumblatt2013,Nam2016,AmazonPLT}.
However, it is often tedious for content providers to quickly pinpoint the location of faults, as this often requires costly human expertise to understand
whether a degradation is due to their own internal infrastructure or to weak Internet peering to specific ISPs, for instance.

To improve on this situation, numerous prior works have proposed to exploit end-user devices and equipment to diagnose on-line incidents~\cite{Dischinger2010,Kreibich2010,Dhawan2012}.
These works adopt two main strategies.
The first is to execute a set of predefined tests, designed by experts in networking, and use outliers to propose a diagnostic~\cite{Sundaresan2011,Sundaresan2013,Dhawan2012,Kreibich2010}.
These tests are very efficient to detect known configuration issues (DNS failures, aggressive firewall, low quality uplink, \dots), but are specific to some technologies (like DNS, TCP or DSL access specific problems~\cite{Jin2010}), and fall short in understanding more distant Internet failures.
The second strategy is to use a shared service status database like downdetector.com to easily discriminate between local or distant fault.
Unfortunately, such services are usually centralized and only offer coarse-grained analysis based on manual flagging:
as an example, if for a given service a large number of reports usually come from Germany, the only thing that can be inferred is that many users of the service in Germany are encountering a fault\ and are willing to share that information.
The precise root cause location is still to be determined.
While the above solutions either provide important insights on service availability or focus on important types of faults, they cover only a small and specific part of the possible root causes for many online services and fail at offering generic internet-scale root cause inference.

To overcome these limitations, we propose \diagnet, a generic and extensible crowd-sourced root cause analysis method based on data collected from user devices.
\diagnet uses browsers to take measurements, and exploits a versatile inference model to diagnose problems proactively before calling the customer support of the ISP or content provider.
Our system relies on a neural network for root cause inference model that does not make any assumption on the underlying network topology and can ingest new types of network measurements without the need for retraining.
\diagnet is based on a set of \emph{landmark servers} that act as reference points for measures. We assume these servers are opportunistically deployed over diverse parts of the Internet independently from any network operator.
\diagnet leverages attention mechanisms, a technique to highlight the input features that were relevant for a particular classification result, in combination with non-overlapping convolutional kernels and pooling mechanisms, borrowing and extending state-of-the art concepts from the image processing community~\cite{LeCun1989,Lin2013,Simonyan2014}.
Doing so, \diagnet is the first Internet-scale network-diagnostic solution that can infer root causes it never encountered before, and can easily be adapted to different types of online services with very little retraining, while only requiring lightweight, easy to obtain user-side measurements.
The principles behind \diagnet are further not limited to end-user problems, and generalize easily beyond Browser-based services, to distributed B2B APIs.

Our main contributions are as follows:

\begin{itemize}
  \item We define a set of properties that must be satisfied for end-user root cause analysis in today's Internet, with no external information on the network topology or inter-services dependencies.
  \item We propose a simple root cause analysis architecture based only on measurements from end-users devices and a dynamic set of landmark servers. While in our implementation and evaluation we chose to focus on measurements available within a browser, any client metric could be exploited by the proposed architecture and models.
  \item We build \diagnet, a root cause inference model that can handle an extensible set of network measurements and therefore pinpoint locations it never encountered before. The proposed model introduces new types of convolutional layers as well as attention mechanisms, which make the model generic and extensible.
  \item We evaluate our proposal on mock-up online services and clients deployed in 10 world regions and relying on multiple cloud providers with various dependencies between services.
    We compare \diagnet against simpler, yet recognized inference methods and show that it consistently overperforms its competitors in a dynamic context,
    that is typical of today's Internet services, while delivering close to ideal performances in a static setting.
    More specifically, \diagnet is able to pinpoint root causes with a Recall@1 of 73.9\%, including non-trained root-causes.
\end{itemize}

The remainder of this paper is structured as follows.
We start by specifying our problem and the set of required properties for \diagnet in~\autoref{sec:problem}.
In~\autoref{sec:approach:diagnet}, we dive in the internals of our proposal, from the architecture overview to the predictions fine-tuning.
Based on a geo-distributed collection of metrics, we propose in~\autoref{sec:evaluation} an extensive evaluation of \diagnet, alongside with two baseline proposals for comparison.
We present related work in~\autoref{sec:related}, and~\autoref{sec:conclusion} concludes this document.

\section{Problem statement and goals}
\label{sec:problem}

After a brief overview of our system's model, we introduce in this section three key properties that we argue are required to design a root cause inference method that is generic and can work at Internet-scale: \emph{network topology agnosticism}, \emph{location agnosticism}, and \emph{root cause extensibility}.

\subsection{System overview}
\label{sec:sysmodel}

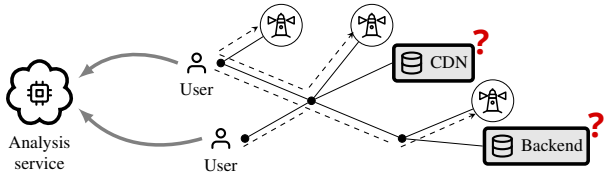
\begin{figure}
  \centering
  \resizebox{0.5\textwidth}{!}{%
    \input{figures_overview.tex}
  }
  \caption{Toy example of a topology for an online-service relying on a CDN and a backend.
  Users can evaluate links (solid lines) by actively probing landmarks \landmarkIcon{} (dashed lines).
  Probes are sent to a root cause analysis service, which builds and shares the root cause inference model.}\label{fig:landmarks}
\end{figure}

Our vision is that of a \emph{central root cause analysis service} that is reachable from any end-user device (also termed \emph{client}).
This service continuously processes measures provided by a subset of clients to maintain a model of the network. This crowd-sourced model is then able to diagnose failures of online services consumed by end-users.
We focus in this paper on the design and construction of this central model, but leave the implementation details of the crowd-sourcing mechanisms that are necessary to aggregate individual measurements to future work.
Clients produce measurements by actively probing \emph{landmark servers} (see~\autoref{fig:landmarks}), i.e.\ stateless public HTTP services that can be provided by different ISPs or third parties.
The global network of Speedtest servers~\cite{SpeedtestNetwork} is an example of practical public landmark servers deployment.
More specifically, we leverage modern web browser capabilities to fetch TCP statistics, latency and bandwidth information from these landmark servers,
to which we add some \emph{local system features} (e.g.\ client CPU and memory load) measured on the client itself.
Within a browser this can either be implemented as a JavaScript that is fetched when accessing the online service\ (the solution we have used for our prototype), or as a browser extension.
While we wanted to keep a very simple and restricted set of metrics to bootstrap our work, it is absolutely possible to add more specific measures as additional inputs.

The measures collected by a client $i$ form a vector of $m$ measures $\mathbf{x}_{i} = (x_{i,j})_{1\leq j\leq m}\in \mathbb{R}^m$ .
They constitute the \emph{features} that are fed into the root cause analysis service. (In the following we use the terms \emph{measure} and \emph{feature} interchangeably.)
We assume clients also collect the Quality of Experience (QoE) perceived by their users through a binary indicator, that records whether a user is experiencing a problem or not for a given service.
This QoE information might be manually provided by users, or automatically estimated. It can be as simple as a page load time or can rely on a method that calculates it~\cite{DaHora2018}.
From that data, and assuming that a user is encountering some QoE degradation, our ambition is to propose a ranked list of probable root causes that \emph{explain} this degradation.
We design our root causes to be the combination between a possibly coarse-grained \emph{location} and a \emph{fault family}, e.g\ ``abnormal latency within the AWS US east coast region'' or ``high jitter within local WiFi connection''.

Our root cause analysis should rank the probable explanation by decreasing probability while ensuring their \emph{usefulness} for the end-user, informally defined as the \emph{additional intelligence given by that feature to locate and understand a specific QoE degradation}.
In our system model, each input feature is representative of a root cause: as an example, if a user encounters a QoE degradation while accessing a video streaming service, and the download bandwidth to a landmark located in Spain abruptly decreases, it makes sense to pinpoint the feature ``Spain landmark download'' as probable root cause.
However, if other landmarks suffer from the same bandwidth degradation, but the local computing load of the client also reported an increase, the ``local CPU load'' feature shall be marked as the most probable root cause.
Providing that many features are available, this makes our model very expressive without the need for manual expert annotation.
This analysis is very different from the problem of \emph{feature selection}, where a restricted set of features are selected for model training.
In our case, a useful feature will help understanding and troubleshooting a QoE degradation; as such usefulness must be computed anew for \emph{each diagnostic}.
We denote the usefulness of feature $j$ in sample $i$ by $\gamma_{i,j}$.

\subsection{Network topology agnosticism}

Internet-services rely on a wide variety of systems, sub-services, and networks to function properly.
This includes data centers, cloud-providers, content delivery networks (CDN), along with a range of autonomous systems and operators networks.
The underlying network architectures and topologies of all these systems are complex, continuously evolving and often unknown.
We argue therefore that an Internet-scale root cause analysis method should not make any assumptions on the hidden network topology, a property that we call \emph{network topology agnosticism}.
This largely departs from common root cause analysis that relies on network tomography~\cite{Rubenstein2002,Duffield2006,Dhamdhere2007,Zhao2009} or bespoke methods for data centers~\cite{Arzani2018,Tan2019} and Software-Defined-Networking~\cite{Tammana2018,Ke2018}.

Root cause analysis requires however some location information to pinpoint the area (e.g. cloud region, point of presence, autonomous system) in which a root cause is likely to be. \diagnet relies on landmark servers to provide this location information while eschewing a precise knowledge of the underlying network.
Landmark servers are easy to deploy, cheap to run and maintain, and can provide a good overview of the network health provided they are present in multiple and diverse vantage points.
The intuition is that if there is a sufficiently wide deployment of landmarks, some of these landmarks will be located in the vicinity of targeted services, or in the path towards them, thus offering telling clues regarding the location and family of the incident impacting a user.

Of course, it is probable that some features vary greatly without any impact on the target service.
By understanding the relations between features and service's performance, a model could infer useful glimpses of information about a network's internal architecture, without exhaustive and costly modeling of the full network topology.
This type of analysis shall be largely sufficient for a user to \emph{understand} the kind of encountered failure, and eventually for an operator to start a deeper investigation with good hints from its customer's devices: this is a first light in the dark for accurate root cause analysis from end devices.

\subsection{Location agnosticism}

Historically, crowd-sourced diagnostic tools have exploited the precise location of every client~\cite{Padmanabhan2005}, both at a geographical (from neighborhood to country) and topological level (from subnet to ISP), to pinpoint failures accurately.
However, obtaining such detailed data from every end-device is often difficult and even undesirable: users might refuse to share their precise location out of privacy concerns, while IP addresses are becoming increasingly difficult to locate due to voluntary obfuscation, carrier-grade NATs, roaming and ongoing deployment of IPv6.

In \diagnet, we propose to circumvent the need for location altogether, and argue instead that a root cause analysis model should be \emph{location-agnostic}:
the same single model should apply to every participant in a crowd-sourcing network.
However, we believe it is acceptable to have distinct models for distinct services, since they are definitely less numerous than possible client profiles while being possibly very diverse.
We show in \autoref{sec:approach:diagnet} how this level of expressiveness can be achieved by revisiting multi-layer perceptrons and convolutions in the context of network diagnosis.

\subsection{The need for extensibility}

In our system model, each feature is representative of a root cause.
In practice, each client takes a number of active network-based measurements from available landmarks, which are then fed along with local metrics into the \diagnet inference model.
Many factors can however alter the \emph{availability} of these landmarks (for instance failures, maintenance, network partition or saturated capacity).
Conversely, if the system contains a very high number of landmarks, individual clients can not be expected to probe every landmark in order to keep overheads low.
As an extreme example, it would require at least 94 000 landmark servers to cover every autonomous system\footnote{Data from Regional Internet Registries as of \formatdate{01}{01}{2020}}, a number clearly too high for exhaustive probing.

To address these issues of scale and dynamicity, we require our generic root cause analysis system to be \emph{extensible}: trained models should be able to consume measurements from a varying number of landmarks, depending on their availability at a given time.
This critical property allows for easy maintenance of the landmarks fleet, that should be as large and diverse as possible.
Since the location of a plausible root cause is directly inferred from landmarks, the
better landmarks cover the Internet
the more precise the resulting inference can be expected to be.
A \emph{root cause extensible} model should still however provide accurate results even when only a subset of landmarks are available.
This implies a number of critical choices in the design of our proposal to avoid frequent model retraining.

\section{\diagnet}
\label{sec:approach:diagnet}
Similarly to most statistical classification models, \diagnet takes as input a vector $\mathbf{x}_{i} \in \mathbb{R}^m$ of features measured by a client $\mathsf{c}_i$ (a ``sample'') and outputs a probability vector of likely classes (the predicted root causes).
Contrary to a typical statistical classifier, however, \diagnet can adapt without retraining to a variable number of input features $m$ (provided by a varying number of landmarks), while supporting a dynamic number of diagnostic classes.

\subsection{General architecture}

Our strategy consists in building models that in a first phase only predict the \emph{family} of encountered faults (if any) without any information on the location or on the root cause features (what we call a \emph{coarse prediction} in the following).
The number of fault families $c$ is fixed, and the resulting prediction is a small-size vector $\mathbf{y} \in [0, 1]^c$ of probabilities.
We arbitrary selected the following set of common families: \emph{nominal} (non-faulty); \emph{uplink latency} for gateway malfunction; \emph{remote link latency}, \emph{link jitter}, \emph{link loss} and \emph{link download bandwidth} for end-to-end issues not related to the local uplink; and \emph{local load} for client device overload.
By keeping the coarse dimension low and constant ($c \ll m$), we build accurate models without an excessive number of samples.

In a second phase, \diagnet uses the coarse prediction vector $\mathbf{y} \in [0, 1]^c$ to return to the input feature space of dimension $m$ and locate the fault, in effect equating the final predicted classes with the space of input features.
We can use any attention mechanism in that step: such mechanisms infer the \emph{weight} of each input feature in the coarse model's prediction, often without any additional training.

The global architecture of \diagnet is depicted in~\autoref{fig:network}.
The coarse prediction phase involves the steps of \circled{1} separating landmark features from local features, \circled{2} processing the landmark features with a specific type of \emph{convolutional neural network}  (detailed in~\autoref{sec:neuralnet}), \circled{3}\circled{4} processing all features with a fully-connected network and obtaining the final coarse prediction  (detailed in~\autoref{sec:servicelayer}).
The second phase involves the step of \circled{5} returning back to the input features via attention mechanisms (\autoref{sec:attention}).

\subsection{Non-overlapping convolutions with pooling}\label{sec:neuralnet}

In image analysis, convolutional neural networks~\cite{LeCun1989} have been used with considerable success to classify images.
Their convolutional layers extract patterns over multiple pixels by applying small filters over each pixel and its neighbors.
We borrowed this idea of \emph{pattern extraction} to extract \emph{common patterns} between different landmarks, with some differences.

First, in contrast with image pixels, we want to combine measures of different nature (linked to ``fault families'', such as latency and bandwidth).
For a landmark $\lambda$, and a client $\mathsf{c}_i$, we note $\mathbf{x}_{i}[\lambda] \in \mathbb{R}^k$ the vector of measures (e.g. RTT, throughput) recorded by $\mathsf{c}_i$ w.r.t to the landmark $\lambda$.
For example, $x_{i,1}[\lambda]$ might store the RTT from $\mathsf{c}_i$ to $\lambda$, and $x_{i,2}[\lambda]$ the throughput.
In this first phase, we seek to extract recurring patterns from each landmark in isolation.
To this aim, we apply a set of $f$ \emph{non-overlapping} convolutions to each client/landmark measure vector $\mathbf{x}_{i}[\lambda]$. These convolutions are captured by a kernel $\mathbf{K} \in \mathbb{R}^{f\times{}k}$ and bias $\mathbf{b} \in \mathbb{R}^{f}$. Formally:
$$
\forall \lambda \in \{1, \dots, \ell\}, \mathbf{F}[\lambda] = \mathbf{K} . \mathbf{x}_{i}[\lambda] + \mathbf{b}.
$$

\newcommand{\pool}{\operatornamewithlimits{\mathlarger{\mathlarger{\mathlarger{\Omega}}}}}

At this stage, the $\ell\times k$ landmark features have been projected into a new feature space of dimension $f$ (the number of filters).
Since the $\mathbf{K}$ and $\mathbf{b}$ parameters are shared for every landmark, we believe that \emph{common patterns between landmarks} are learned: our model shall hopefully extract useful information about the underlying network architecture.
Nevertheless, it is still required to return a vector which size is independent of the number of available landmarks.
We thus leverage global pooling layers~\cite{Lin2013,He2015}, a popular mechanism to support variable-size inputs and ensure good generalization in image analysis.
In our case, we apply a global function $\Omega$ on every landmark's convolution feature element-wise:
$$
\mathbf{F} = \pool_{\lambda = 1}^{\ell} \mathbf{K} . \mathbf{x}_{i}[\lambda] + \mathbf{b}, \mathbf{F} \in \mathbb{R}^f
$$

\begin{figure}[t]
  \centering
  \resizebox{0.45\textwidth}{!}{%
    \input{figures_landpooling.tex}
  }
  \caption{Graphical view of the proposed non-overlapping convolutional layer with pooling (LandPooling).
  For each landmark $\lambda$, the $k$ features of that landmark $\mathbf{x}_{i}[\lambda]$ are transformed to a new feature space $\mathbf{F}[\lambda]$ of size $f$ through a shared kernel $\mathbf{K}$.
  To return a fixed-size output of size $f$, the results for the $\ell$ landmarks are combined through a global $\Omega$ function, such as maximum, average or others.
  }\label{fig:landpooling}
\end{figure}
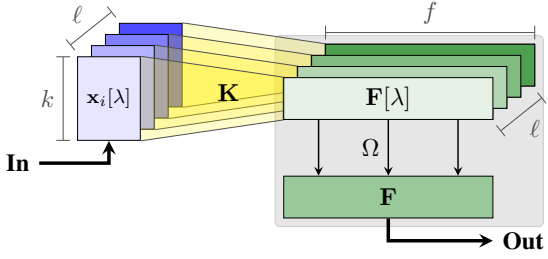

We define this process as a new kind of neural network layer, and call it ``LandPooling'' by reference to landmarks.
An illustration of this landmark-flattening process is depicted in~\autoref{fig:landpooling}.
We note that any commutative function that can be applied with a generic number of arguments can be chosen for $\Omega$.
The hyperparameters and global functions we used in our implementation of \diagnet are listed in~\autoref{table:hyperparameters}.

\begin{figure}[t]
  \centering
  \resizebox{0.5\textwidth}{!}{%
    \input{figures_network.tex}
  }
  \caption{Architecture of \diagnet.
  \circled{1} Landmark features are first separated from local features and  \circled{2} fed in the LandPooling layer with multiple parallel global pooling operations.
  \circled{3} A hidden fully-connected layer is applied after concatenating the LandPooling output with local features.
  \circled{4} The coarse fault prediction is obtained by applying a series of non-linearities.
  \circled{5} Finally, an attention model is applied on the coarse prediction to return to the feature space and propose a fine-grained fault localization.}\label{fig:network}
\end{figure}
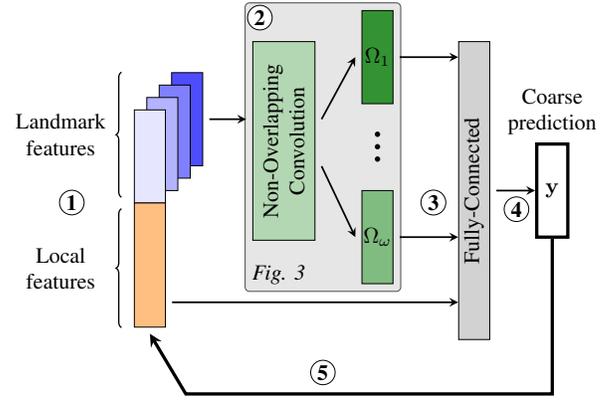

\begin{table}[tb]
  \renewcommand{\arraystretch}{0.9}
  \caption{Notations and hyperparameters}\label{table:hyperparameters}
  \centering
  \begin{tabular}{ll}
  \toprule
  $\ell$ & Total number of landmarks (10) \\
  $f$    & Number of convolutional filters (24) \\
  $k$    & Number of features per landmark (5) \\
  $m$    & Number of features per sample ($\ell\times{}k + \text{local features} = 55$) \\
  $c$    & Number of coarse fault families (7) \\
  $\Omega$ & Global pooling operations (min, max, avg, variance, p10, \dots, p90) \\
  $\mathbf{x}_i$ & $= (x_{i,j})_{1\leq j\leq m}$, input sample of client $\mathsf{c}_i$ \\
  $\mathbf{y}_i$ & $= (y_{i,j})_{1\leq j\leq c}$, coarse predictions for client $\mathsf{c}_i$ \\
  $\mathbf{\hat\gamma}_i$ & $= (\hat\gamma_{i,j})_{1\leq j\leq m}$, predicted features usefulness for $\mathsf{c}_i$ \\
  \midrule
  & Fully-Connected layers: 2 hidden layers $(512 \times 1), (128 \times 1)$ \\
  & Learning algorithm: SGD with Nesterov momentum \\
  & (learning rate = 0.05, decay = 0.001) \\
  \bottomrule
  \end{tabular}
\end{table}

\subsection{Tailoring to specific services}\label{sec:servicelayer}

Similarly to classical classification tasks relying on convolutional architectures, we propose to add a multi-layer perceptron after the LandPooling mechanism presented in the previous subsection.
The main purpose of these additional layers is to increase the expressivity of \diagnet, by permitting a non-linear combination of the results of the global pooling and the local features resulting in coarse fault predictions.
As illustrated in~\autoref{fig:network}, this perceptron (also called ``Fully-Connected layers'') accepts multiple inputs: the global pooling functions $\Omega_{1}, \dots, \Omega_{\omega}$ along with the ``local features'' that are independent of available landmarks.
This additional expressivity is necessary to model the dependencies between services and input features.
By default, \diagnet uses one single \emph{general} set of final fully-connected layers to diagnose multiple services.

However, such \emph{general} model could demonstrate variable performance if the set of monitored services is very diverse:
not all Internet services have the same network requirements and dependencies.
For example, while the latency is absolutely critical in multiplayer games, it might intuitively not be the case for video streaming systems where the bandwidth is usually the bottleneck.
It is thus possible to build one \emph{specialized} \diagnet model per service to improve its accuracy, by learning a dedicated set of fully-connected layers for that service.
We detail and evaluate this property in~\autoref{sec:transfer}.

\subsection{Fine-grained inference via attention mechanisms}\label{sec:attention}

To offer a fully extensible model, we need a mechanism to evaluate the \emph{importance} of each input feature (each possible root cause) in the coarse-grained fault prediction.
There exist techniques to directly evaluate such importance in simple models (e.g. decision trees), but it is well-known that this kind of \emph{attention evaluation} is non-trivial for neural networks.
While some generic techniques are applicable to any black-box model including ours~\cite{Ribeiro2016}, we instead propose to compute the \emph{gradients} of the coarse predictions with respect to the input features.
This method has already been tested in image analysis with great success~\cite{Simonyan2014,Selvaraju2017}, and takes advantage of the fact that we can observe the internal weights and architecture of the coarse model (white-box setup).
Given a coarse prediction $\mathbf{y} = (y_{j})_{1\leq j\leq c} \in \mathbb{R}^c$ (step \circled{4} of~\autoref{fig:network}), we first compute the \emph{ideal label vector} $\mathbf{y^\star}$ that would have been given during the training for the input sample.\footnote{For readability, and without ambiguity since we are now working on a single sample $i$, we removed the $i$ indices of all notations.}

$$
\forall j \in \{1, \dots,  c\}, y^\star_j = \left\{\begin{matrix*}[l]
1 & \text{if} \max(\mathbf{y}) = y_{j} \\
0 & \text{otherwise}
\end{matrix*}\right.
$$

We define
$
L^\star(\mathbf{y})
= - \sum_{j = 1}^{c} y^\star_j\log{}y_j
= - \log{}y_{\argmax(\mathbf{y})}
$ the cross-entropy loss that is minimal for the ideal label vector.
By applying a single backpropagation step as done during the training phase, and thanks to the complete knowledge of the coarse model architecture,
we can compute the \emph{gradient} of this loss function with respect to the input features.
We make the assumption that each partial derivative $\nabla_{j} = \frac{\partial L^\star}{\partial x_{j}}$ represents the \emph{usefulness} of each feature $j$, and can be normalized according to the absolute value of $\nabla_{j}$ to account for both positive and negative derivatives.

\begin{equation}
    \hat{\gamma}_{i,j} = \frac{|\nabla_{j}|}{\sum_{k}{|\nabla_{k}|}}
    \label{eq:attention}
\end{equation}

In our early experiments, we observed that the attention mechanism (\autoref{eq:attention}) used alone as a predictor of root causes gave very inaccurate results.
This is because a pure gradient-based backpropagation does not fully exploit the information provided by the multi-layer perceptron (\circled{4} in \autoref{fig:network}).
To overcome this problem, we give a bonus to the most relevant root causes that belong to the same family (e.g. latency, bandwidth) as the most probable coarse cause returned by the coarse prediction.
For instance, if the model predicts a \emph{remote link latency} problem, we use this hint to increase the predicted usefulness of every latency-related feature while penalizing other features.
The weighting mechanism is shown in~\autoref{alg:tuning}.

{\scriptsize
\input{tuning}
}

Given a coarse prediction vector $\mathbf{y}$, the algorithm first selects a set of features $p$ related to the most significant class in $\mathbf{y}$ (in practice of the same family) at~\autoref{tuning:relation}.
In our implementation, we manually assign each feature to a coarse class.
Then, a ratio $w$ is computed between the model's confidence in its coarse prediction and the sum $s$ of related features' usefulness (\autoref{tuning:weight}).
The tuned $\hat\gamma'$ are computed in~\autoref{tuning:in} and~\autoref{tuning:out}.
By construction,~\autoref{alg:tuning} always returns a normalized vector.

\subsection{Ensemble model averaging}\label{sec:averaging}

The architecture of~\autoref{fig:network} is designed to naturally extend to new landmarks without retraining. As a result, however, it loses information compared to more direct methods such as random forests. To further boost our solution, and reap the benefits of both worlds, we use \emph{ensemble model averaging} as a last optimization step, a popular method to combine multiple specialized models~\cite{Madigan1996}.
We average the tuned attention predictions with another prediction from an \emph{auxiliary} model, designed to be simpler and specialized in \emph{known root causes}.
We chose a random forest approach as our auxiliary model, and give more insights about this choice in the next section.

We briefly formalize this last optimization.
Let $\mathcal{U}$ be the set of unknown landmark's features, not seen during training.
Let $\hat\gamma'$ and $\hat\alpha$ be the prediction obtained from the tuned attention mechanism and the auxiliary model, respectively. 
We define $w_\mathcal{U}$, the probability that the root cause is explained by an unknown landmark's features, as predicted by the tuned attention mechanism.
Since $w_\mathcal{U} \in [0,1]$ by definition, the final prediction of \diagnet after model averaging is given by

$$
w_\mathcal{U}~\hat\gamma'~+~(1 - w_\mathcal{U})~\hat\alpha~~~~\text{with}~~~~w_\mathcal{U} = \sum_{j \in \mathcal{U}} \hat\gamma_{j}'
$$

\section{Evaluation}\label{sec:evaluation}

To evaluate \diagnet, we deployed a multi-cloud geo-distributed network of clients, online services, and landmark servers.
In this section, we present our methodology and introduce baselines offering similar properties as \diagnet.

\subsection{Experimental setup}

\begin{figure}
  \centering
  \resizebox{0.5\textwidth}{!}{%
    \input{figures_deployment.tex}
  }
  \caption{Locations of landmarks and services in our multi-cloud experimental deployment.
  We emulated clients in every location (region).}\label{fig:world}
\end{figure}
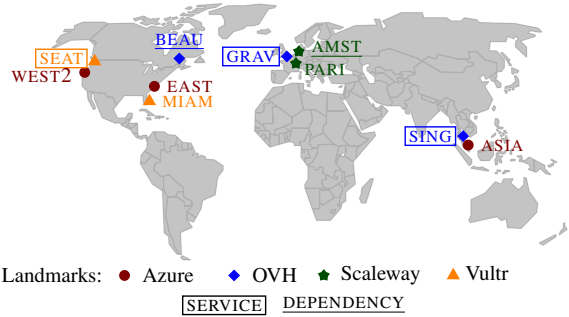

\textbf{Deployment.}
In order to train and evaluate the root cause analysis models, we deploy one landmark and multiple clients in each of the ten regions listed in~\autoref{fig:world} and~\autoref{table:services}.
Three of these regions (\textsc{grav, seat, sing}) also host mock-up online services to evaluate the QoE with diverse setups.
Some services only require a single HTML file, while others download resources from distant regions. (Recall that the nature of individual services, and hence the relations between regions and services are hidden during model training.)
Region locations are chosen to benefit from both the diversity of a worldwide multi-cloud deployment and the proximity of co-located regions for fault localization.
At the time of writing, our experimental pipeline was made of roughly 5000 lines of Python and Go code.
We used Tensorflow 1.13.1 as our machine learning framework.

\textbf{Landmark features.}
Live network metrics (\autoref{table:sample}) are obtained by querying each landmark through HTTPS endpoints.
This choice allows end-users to access landmark features via their web browsers only.
To estimate download and upload bandwidths, we measure the time required for large GET and POST HTTP requests.
We avoided the classic overhead of HTTP requests for Round Trip Time (RTT) estimation by upgrading the connection to WebSocket.
Finally, we use the \texttt{getsockopt} linux syscall on each landmark server to make raw TCP statistics available to landmarks' clients.
We mainly extract the ratio of reordered and retransmitted packets from these statistics.
(For completeness, we add that we used the BBR congestion algorithm for every communication.)

\textbf{Methodology.}
Clients periodically fetch network features from landmarks and visit mockup services to evaluate their QoE from performance timings, both operations using a headless Chromium process.
We inject artificial network faults using Linux \texttt{tc} Network Emulator rules~\cite{linuxtc}, a realistic and popular emulation method for reproducible experiments.
QoE information was then used to flag samples as ``nominal'' or ``faulty'' with the (known) root-cause ground-truth as class label for model training.

\textbf{Root cause extensibility.}
We trained and tested root cause models on two different sets of landmarks to assess the extensibility capabilities of our approach.
For all experiments in this paper, three landmarks were ``hidden'' during training: \textsc{east}, \textsc{grav} and \textsc{seat}, named \emph{new landmarks} as opposed to \emph{known landmarks} (the remaining seven).
We chose these landmarks due to their immediate proximity to the mock-up services and several injected faults, and limited the availability of their features to model evaluation only.
In doing so, we reduced the quality of the measures available to training, and made faults located close to the hidden landmarks particularly hard to detect, as neither these faults, nor the measures they impact most are used to train the models.

\textbf{Dataset.}
We ran our experiment during the last two weeks of December 2019, using different hours of day and days of week to ensure large coverage of traffic and congestion patterns between cloud providers.
We injected the following 6 families of faults in different regions, leading to 24 different fault scenarios:
\begin{itemize}
  \item Download bandwidth shaping (capped at 8 Mbits/sec),
  \item Additional service latency (50 msec),
  \item Additional gateway latency (50 msec),
  \item Additional jitter (up to 100 msec),
  \item Increased packet loss (8\%),
  \item Large CPU stress (this is critical for headless Chromium).
\end{itemize}
Faults were uniformly distributed between regions and families to avoid bias towards more frequent root causes.
In many cases, we observed that the QoE was not degraded despite the injected fault(s).
For instance, the QoE of a small HTML website was not affected by shaped bandwidth or CPU stress.
In this case we flag the samples as ``nominal''.
213 000 of ``nominal'' samples along with 30 000 ``faulty'' samples were collected during our experiment.
80\% of each kind of samples were used for training, while the other 20\% were reserved for testing.

\begin{table}[tb]
  \renewcommand{\arraystretch}{0.9}
  \caption{Regions and online services used in experiments.}\label{table:services}
  \centering
  \begin{tabular}{ll}
    \toprule
    \textbf{Provider} & \textbf{Regions} \\
    \midrule
    Azure         & US (\textsc{east}, \textsc{west2}), Singapore (\textsc{asia}) \\
    OVH           & Beauharnois (\textsc{beau}), Gravelines (\textsc{grav}) \\
                  & Singapore (\textsc{sing}) \\
    Scaleway      & Paris (\textsc{pari}), Amsterdam (\textsc{amst}) \\
    Vultr         & Miami (\textsc{miam}), Seattle (\textsc{seat}) \\
    \midrule
    \textbf{Service} & \textbf{Description} \\
    \midrule
    1. single        & Static HTML page with no dependency \\
    2. script.far    & Requires a JS file in \textsc{beau} \\
    3. script.cdn    & Requires a JS file from the nearest region \\
    4. image.local   & Loads a 5MB image from the same server \\
                  & (using the same HTTP connection) \\
    5. image.far     & Loads a 5MB image from from \textsc{beau} \\
    6. image.cdn     & Loads a 5MB image from the nearest region \\
    \bottomrule
  \end{tabular}
\end{table}

\begin{table}[tb]
  \renewcommand{\arraystretch}{0.9}
  \caption{Features collected during our experiments.\hspace{\textwidth}
  This set of features can easily be extended.}\label{table:sample}
  \centering
  \begin{tabular}{ll}
    \toprule
    \textbf{Type} & \textbf{Features} \\
    \midrule
      Service QoE & Timings from \texttt{window.performance} \\
                  & JavaScript (full page and per resource) \\
    \midrule
      Landmarks (x10) & Download, Upload, Round-trip time, \\
                      & number of TCP reorderings and retransmissions \\
    \midrule
      Local & Total and available memory, \\
            & disk and CPU load, \\
            & round-trip time to gateway \\
    \bottomrule
  \end{tabular}
\end{table}

\subsection{Comparison baselines}

To the best of our knowledge, we are the first to propose a root cause analysis method that is \emph{extensible}, both in the features and causes dimensions, and that does not exploit additional information on network dependencies.
We propose two baselines that use common classification models and offer the same three key properties as \diagnet, namely location and topology agnosticism, along with root cause extensibility.

\textbf{Extensible Random Forest Classifier.}
A random forest ensemble classifier is built by constructing a large set of small decision trees.
The final classification is done through a majority vote on trees outcomes.
This method is well-understood and has been previously used in failure classification in NetPoirot~\cite{Arzani2016}, showing great stability and accuracy in the face of diverse machines.
To train an extensible random forest, we naively set the features dimension to the maximum possible size, and we set to zero the missing landmarks values in each sample.
We also add a special ``unknown'' output class, selected when the given sample is classified as ``nominal''.
We evenly redistribute the score obtained for this special class to every other class: this allow non-trained faults to have a non-null score in the final prediction.
This model is used as-is in the ensemble averaging optimization presented in~\autoref{sec:averaging}.

\textbf{Extensible Naive Bayes Classifier.}
We propose another approach for extensibility, based on the merger of several probability distributions.
Using Bayes theorem and making the ``naive'' assumption that the value of one feature is not dependent from other features, it is easy to compute the posterior probability that one sample belongs to a class $C_k$ given the estimated prior and likelihood probabilities.
\autoref{eq:naivebayes} presents the application of the Bayes theorem for classification.

\begin{gather}
P(C_{k} \mid \mathbf{x}_{i}) \propto P(C_{k}) \prod_{j = 1}^{m} P(x_{i,j} \mid C_{k}) \label{eq:naivebayes}
\end{gather}

To add a basic support for model extensibility, we adapt the classic model in the following way:
\begin{itemize}

\item First, it is highly probable that one particular root cause $C_k$ has not been seen during the training phase, and the prior probability for class $k$ is unknown.
Thus, we define the prior probability of each class $C_k$ as $P(C_k) = 1$ for every root cause.
This also has the positive side-effect of canceling bias with unbalanced datasets.

\item Second, we use a Kernel Density Estimation (KDE)~\cite{Rosenblatt1956} function to construct the likelihood probabilities $P(x_{i,j} \mid C_{k})$.
In contrast with the more common Gaussian model, the KDE increases the expressivity of this baseline model.

\item Finally, we build \emph{generic aggregate likelihoods} for unknown features or new classes.
For each measure family $t$ collected in landmarks (such as uplink latency or download bandwidth), we build a \emph{generic} likelihood $P(x_{i,t} \mid C_t)$.
This generic likelihood is the union KDE of the measures \emph{for every landmark available during training},
and becomes the default when no specific likelihood is available for a given feature or class.
\end{itemize}

\subsection{Recall evaluation}

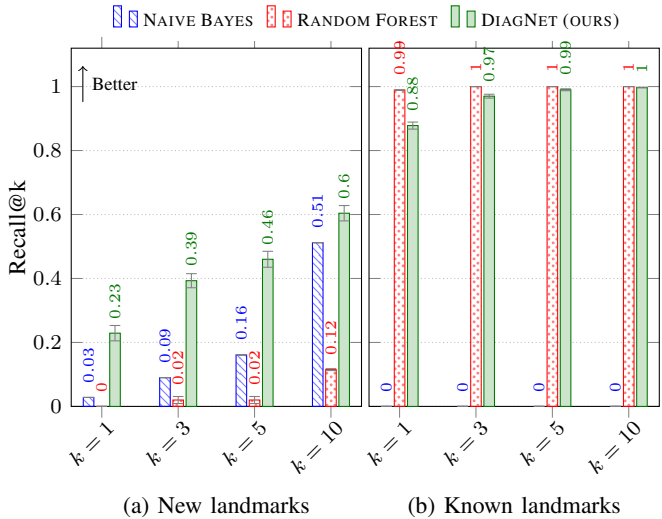
\begin{figure}
	\centering
	\input{plots_recalls.tex}
	\caption{Evaluation of models recalls for failures near new and known landmarks, for different levels of recall $k$. \diagnet consistently overperforms its competitors on new landmarks, while delivering close to ideal performances on known ones. By comparison \textsc{Random Forest} works perfectly for known landmarks, but degrades starkly on new ones, while \textsc{Naive Bayes} offers reasonable performance with new landmarks, but is lost on old ones.
	}\label{fig:recalls}
\end{figure}

The final goal of root cause analysis is to return a \emph{ranked list of probable causes} to users and operators.
We propose to leverage the Recall@k metric for model evaluation: for a set of known real causes and a ranking method, the Recall@k is the number of correctly predicted causes \emph{within the first $k \geq 1$ causes} divided by the total number of causes.
A high recall would demonstrate that a method of ranking (model) can be useful to users, being able to quickly pinpoint the real root cause of a QoE degradation among a set of possible causes.
In our setup, we argue that it is acceptable to return the expected cause within the first $k \leq 5$ predictions from 55 possible root causes.

\autoref{fig:recalls} shows the Recall@k for two types of fault: faults injected near \emph{new landmarks} in (a), and faults injected near \emph{known landmarks} in (b).
(As a reminder, new landmarks' features are hidden during training.)
\diagnet offers the best recalls for faults near new landmarks (a), thanks to its attention mechanism that fully exploits the information coming from the new features without additional training.
Our proposal also shows close to ideal results for faults injected near known landmarks (b), thanks to the ``hybrid'' mode of operation offered by ensemble averaging (\autoref{sec:averaging}).
The combined Recall@1 for \diagnet (including faults near known and new landmarks) is 73.9\%, a very good score given the high number of probable root causes.
By contrast, \textsc{Random Forest} works perfectly for known landmarks, but its recall degrades dramatically in the case of new landmarks.
This is understandable, as the described extensible random forest model essentially gives \emph{completely random predictions} in this second case.
By contrast, \textsc{Naive Bayes} shows extremely poor results for known landmarks, with its best score reached for high values of $k$ in (a).
This is due to a severe bias towards new features that systematically get high prediction scores \emph{even for known failure types}.

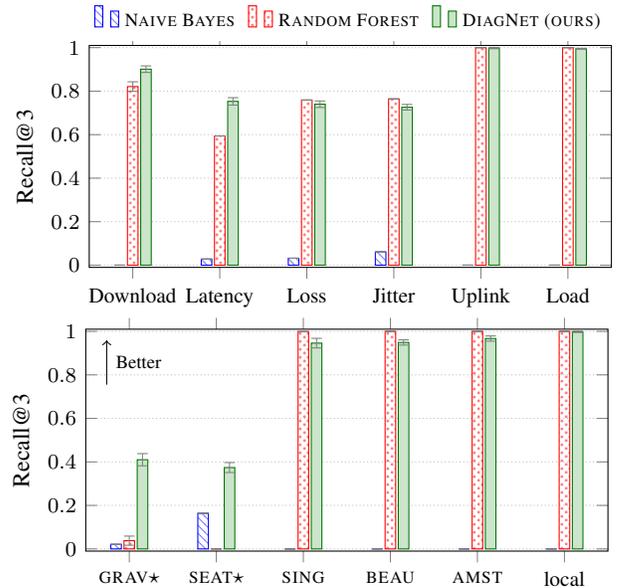
\begin{figure}
  \centering
  \input{plots_recalls_details}
  \caption{Detail of models recalls per fault family (top) and fault region (bottom).
  Regions hidden during training are indicated with a star $\star$.
  Again, \textsc{Random Forest} gives best results for known landmarks, but \diagnet is the only solution able to adapt to the different scenarios, with optimal results for local faults.}\label{fig:recall_details}
\end{figure}

Diving into the details, \autoref{fig:recall_details} presents each recall per family of fault and per location.
We clearly see \textsc{Naive Bayes}'s bias towards some fault families and new landmarks \textsc{grav} and \textsc{seat}.
\diagnet is the only model demonstrating its versatility with good recalls for every family of fault in both known and new landmarks regions.

\subsection{Effect of client diversity}

\begin{figure}
  \centering
  \input{plots_clients}
  \caption{Comparison of models' performance with increasing diversity of clients as the number of regions with active clients.
  \diagnet can scale very well for both known and unknown landmarks, with stable recall. The \textsc{Naive Bayes} approach seems to be optimal at 5 different regions with users, and does not scale with more available regions.}\label{fig:clients}
\end{figure}
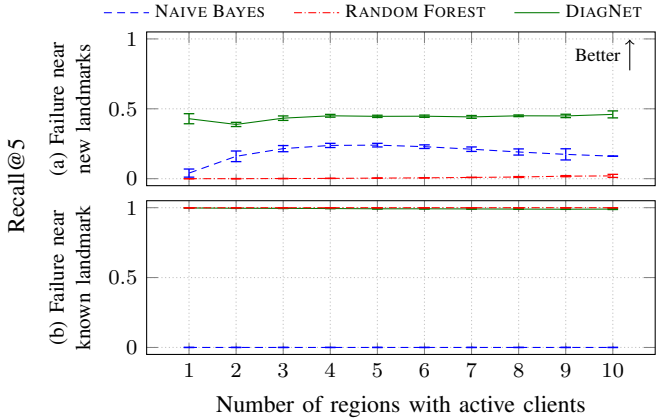

To validate the \emph{location agnosticism} property of \diagnet, we gradually increase the location \emph{diversity} of participating clients.
(Put differently, we vary the \emph{number of regions with active clients submitting samples}.)
The results of that experiment are shown in~\autoref{fig:clients}, with the aggregate Recall@5 for all families of faults near newly-introduced landmarks (a) and near known landmarks (b).
For completeness, we note that we measured the Recall@5 for every possible combination of active clients to eliminate potential discrepancies between configurations.
The key take-away is that \diagnet is able to give the best predictions for all scenarios of client diversity, showing great stability.
Our results hint that \diagnet is truly able to distinguish between dissimilar clients (e.g. clients in America vs. Asia or Europe).

In contrast, the \textsc{Naive Bayes} model prefers to handle few regions at a time.
This is explained by the \emph{KDE merge} process of this baseline: with more diverse clients, merged KDEs are ``flattened'' and converge to uniform distributions, biasing the model towards unknown features as seen in~\autoref{fig:recalls} and~\autoref{fig:recall_details}.
\textsc{Random Forest} is less sensitive to client diversity, with only a slight recall increase in the (a) case, probably due to the growing number of training samples.

\subsection{Training cost of new service models}\label{sec:transfer}

\diagnet\ combines a weighted attention mechanism\ with a non-overlapping convolutional kernel for generalization, and pooling layers for extensibility.
To remove the need for the complete retraining of \diagnet when new online services are being added, we assume
that the weights learned in the non-overlapping convolution are shared between services, as they extract global network features;
and that the final layers of \diagnet capture the behavior of each service.
We now give the details of the \diagnet learning procedure, that has been used in the whole evaluation section and is based on that assumption.
We first train a \emph{general} model on a subset of eight initial services, taking the union of services' problems as the expected model output.
Then, we freeze the weights of the non-overlapping convolution, and optimize the weights of the final layer for each of a set of additional services, not contained in the original set.
This leads to one \emph{specialized} model per additional service.

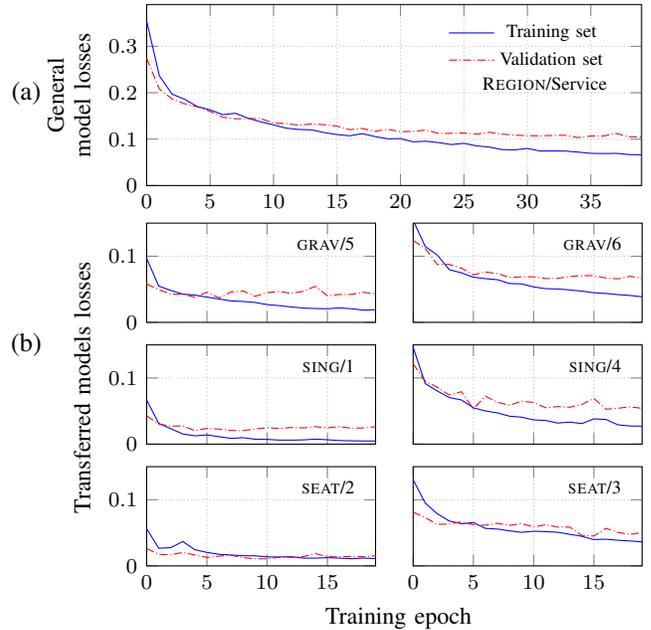
\begin{figure}
  \centering
  \input{plots_history}
  \caption{Evaluation of model transferability: after building a general model on 8 services chosen uniformly at random (a), it is possible to build specialized models for other services while freezing convolutional kernels (b).
  A relatively low loss rate is quickly reached for most services.}\label{fig:learning_history}
\end{figure}

Learning losses on training and validation sets are plotted in~\autoref{fig:learning_history} through learning epochs, for the general model and for a subset of service models.
We consider that the training is done when the validation loss is no longer decreasing (an indication of overfitting).
Although the training time on the general model is higher (around 20 learning epochs), service models converge in less than 5 epochs on average.
This indicates that specialized service models per service are easy to learn once one global model exists.
We note that while the general model can be trained with a subset of services and landmarks, it can later be generalized to more landmarks and services with minimal re-training time.

\section{Related work}\label{sec:related}

Two measurement approaches exist for data collection in root cause analysis solutions: \emph{passive observation} and \emph{active probing}.
Passive measurement relies on existing traffic and does not introduce any overhead and has therefore been previously used in large-scale systems~\cite{Padmanabhan2005,Mysore2014}.
Depending on the setup, different sources of measurements are available such as local system or router insights~\cite{Yan2010,Arzani2016,Roy2017}, request path annotations~\cite{Chen2004b,Fonseca2007}, routing monitors~\cite{Giotsas2017} or more recently from Internet background radiation~\cite{Guillot2019}.
Still, passive observations fall short in low-traffic environments with little information about the underlying network architecture and no control over routing paths~\cite{Fok2013}.
\diagnet leverages active probing to perform accurate root cause analysis from end-devices alone, that have a very narrow view of the underlying network topology.
We note that this method can also be used in conjunction with passive monitoring for bootstrapping a system~\cite{Padmanabhan2005}, or testing hypotheses~\cite{Rish2004}.

Numerous works have been performed in enterprise networks and datacenters, where the full network topology is known (e.g Clos-like topologies in Azure, or SDN-driven networks~\cite{Tammana2018,Ke2018}).
This topology information allows network tomography techniques to be applied~\cite{Castro2004,Rubenstein2002,Duffield2006,Dhamdhere2007,Zhao2009,Ghita2010,Arzani2018,Tan2019}, pinpointing faulty links or components accurately at scale despite complex dependencies between components~\cite{Zhang2018}.

From a metrics perspective, root cause analysis tools are often specialized towards a  set of metrics and thus a narrow set of faults. For example, some focus on poor TCP statistics~\cite{Rubenstein2002,Sundaresan2013,Arzani2016,Arzani2018}, on invalid BGP announcements~\cite{Yan2010,Giotsas2017} or Virtual Hard Drive failures in the recent DeepView work~\cite{Zhang2018}. Netalyzr~\cite{Kreibich2010} and Fathom~\cite{Dhawan2012} collect end-user connectivity statistics to propose automated troubleshooting using predefined expert rules.

By contrast, our work keeps a generic approach and avoids solutions exclusive to known topologies or specific connectivity rules.
Nevertheless, the specific metrics extracted from the aforementioned methods are complementary to \diagnet and could be used as additional input features if available, allowing for narrower and more accurate root cause diagnostics.

Regarding machine learning methods, belief networks have been heavily used in root cause analysis strategies~\cite{Steinder2002,Bahl2007,Kiciman2008} to model the complex dependencies between network components and online services.
However, such methods require many approximations in the modeling and the solving to remain tractable, while being very sensitive to errors in topology identification~\cite{Coates2002,Huang2008}.
Random forest models are also known to perform well in understanding network failures, as demonstrated by NetPoirot~\cite{Arzani2016}.
To the best of our knowledge, \diagnet is the first attempt to model a network with a variable number of input features (landmarks) using convolutional neural networks.
\diagnet relies on the good expressivity of non-linear models to learn features dependencies, and ensures a low training cost by proposing generic models applicable to multiple network configurations and services without complex dependency modeling.

Crowd-sourcing measurements and root cause analysis is a promising approach, where multiple vantage points share their results to better estimate fault root causes~\cite{Clark2003,Wang2004,Kiciman2008,Padmanabhan2005}.
Distributed Hash Tables (DHT) have historically been used for that purpose~\cite{Choffnes2010,Kim2014}, ensuring the scalability of such decentralized systems.
This line of work is complementary to \diagnet: while we currently assume that the analysis process and data collection is handled by a centralized location, DHTs or other distributed system approaches could be used to distribute the root cause analysis service.

\section{Conclusion and future work}\label{sec:conclusion}

Root cause analysis at the scale of the Internet is recognized as a hard problem given the decentralized design of the network.
In this work, we have proposed \diagnet, a generic and extensible crowd-sourced root cause analysis method based on active landmark probing. \diagnet does not depend on prior network topology or service knowledge which makes it practical for end-users that have a very limited view of the Internet topology past their gateway.
The inference model of \diagnet relies on a new type of convolutional network and attention mechanism, along with several optimizations (multi-label score weighting and ensemble model averaging).
While we demonstrated that Random Forest models can be very insightful when diagnosing in a \emph{static} setting
and that Naive Bayesian approaches can also be leveraged for some faults in more dynamic settings,
\diagnet shows good results in \emph{all} scenarios, i.e. it can diagnose local and remote failures in static and dynamic network settings, even with very diverse participating clients  from across the globe.

Our future research is now focusing on deploying and validating \diagnet for a large set of real online services.
There is still a lot of possible improvements in the neural network architecture, for instance by modifying non-linear operations or adding more convolutional or hidden layers.
One avenue for future work would be the integration of historic data in landmark features.
We are also working towards a decentralized version of \diagnet to propose a more scalable and privacy-preserving solution, thanks to recent advances in federated learning and training over encrypted data.

\bibliographystyle{IEEEtran}
\IEEEtriggeratref{40}
\bibliography{sources.bib,web.bib}

\end{document}

%% file: abstract.tex
Diagnosing problems in Internet-scale services remains particularly difficult and costly for both content providers and ISPs.
Because the Internet is decentralized, the cause of such problems might lie anywhere between an end-user’s device and the service datacenters. Further, the set of possible problems and causes is not known in advance, making it impossible in practice to train a classifier with all combinations of problems, causes and locations.

In this paper, we explore how different machine learning techniques can be used for Internet-scale root cause analysis using measurements taken from end-user devices. We show how to build generic models that
(i) are agnostic to the underlying network topology,
(ii) do not require to define the full set of possible causes during training, and
(iii) can be quickly adapted to diagnose new services.
Our solution, \diagnet{}, adapts concepts from image processing research to handle network and system metrics.
We evaluate \diagnet{} with a multi-cloud deployment of online services with injected faults and emulated clients with automated browsers.
We demonstrate promising root cause analysis capabilities, with a recall of 73.9\% including causes only being introduced at inference time.

%% file: figures_overview.tex
\newcommand{\addIcon}[1]{\raisebox{-.3\height}{\includegraphics[height=0.5cm]{figures_#1.pdf}}}

\tikzstyle{dot}=[draw,circle,fill,minimum size=1.5mm,inner sep=0pt]
\tikzstyle{landmark}=[draw,circle,inner sep=1pt]
\tikzstyle{service}=[draw,anchor=west,xshift=-2mm,inner sep=1pt,line width=2pt,inner sep=3pt,fill=black!10,rounded corners=2pt]
\tikzstyle{link}=[draw]
\tikzstyle{probe}=[draw,dashed,>=stealth,semithick,transform canvas={xshift=-1mm,yshift=2mm}]
\tikzstyle{diag}=[draw=black!50,line width=2pt,->,>=latex]

\begin{tikzpicture}[x=1.2cm,y=1.5cm]
  \node (user1) at (-0.4, 0) {\addIcon{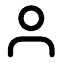}};
  \path (user1.south)+(0,-0.1) node {User};

  \node (user2) at (0, -1) {\addIcon{user}};
  \path (user2.south)+(0,-0.1) node {User};

  \node[dot] (d1) at (0, 0) {};
  \node[dot] (d2) at (1.5, -0.5) {};
  \node[dot] (d3) at (3, -1) {};
  \node[dot] (d4) at (0.4, -1) {};

  \node[landmark] (l1) at (1, 0.5) {\addIcon{lighthouse}};
  \node[landmark] (l2) at (2.5, 0.5) {\addIcon{lighthouse}};
  \node[landmark] (l3) at (4.5, -0.5) {\addIcon{lighthouse}};

  \node[service] (s1) at (3, 0) {\addIcon{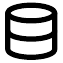} CDN};
  \node[service] (s2) at (4.5, -1.1) {\addIcon{database} Backend};
  \path (s1.north east)+(0.1,0.02) node {\addIcon{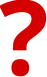}};
  \path (s2.north east)+(0.1,0.02) node {\addIcon{question}};

  \node (cloud) [cloud,draw=black,line width=2pt,aspect=1,cloud puffs=7] at (-3, -0.4) {\addIcon{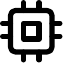}};
  \path (cloud.south)+(0,-0.1) node [text width=2cm,anchor=north,align=center] {Analysis\\service};

  \path[link] (d1) -- (d2) -- (d3) -- (s2.west);
  \path[link] (d1) -- (l1);
  \path[link] (d2) -- (l2);
  \path[link] (d3) -- (l3);
  \path[link] (d2) -- (s1.west);
  \path[link] (d4) -- (d2);

  \draw [diag,bend right] (user1.west) to ($(cloud.east)+(0.1,0.2)$);
  \draw [diag,bend left] (user2.west) to ($(cloud.east)+(0.1,-0.2)$);

  \path[probe,->] (d1) -- (l1);
  \path[probe,->] (d1) -- (d2) -- (l2);
  \path[probe,transform canvas={xshift=-1mm,yshift=-3mm}] (d1) -- (d2) -- (d3);
  \path[probe,->,transform canvas={yshift=-4mm}] (d3) -- (4.3, -0.54);
  \path[probe,transform canvas={yshift=-4mm}] (d4) -- (d2);
\end{tikzpicture}

%% file: figures_network.tex
\definecolor{darkgreen}{RGB}{0,120,0}
\definecolor{ggray}{RGB}{30,30,30}

\tikzstyle{link}=[->,>=stealth,thick]
\tikzstyle{landmark}=[draw, minimum width=0.5cm, minimum height=1.5cm, fill=white, anchor=north]
\tikzstyle{dot}=[draw,shape=circle,minimum size=2pt,inner sep=0,fill=black]
\tikzstyle{circled}=[draw,circle,inner sep=1pt,fill=white]

\begin{tikzpicture}[x=1cm,y=1cm]

  \path node (i1) [landmark, fill=blue!70] {}
    ++(-0.2,-0.2) node (i2) [landmark, fill=blue!50] {}
    ++(-0.2,-0.2) node (i3) [landmark, fill=blue!30] {}
    ++(-0.2,-0.2) node (i4) [landmark,fill=blue!10] {}
    ++(0, -1.5) node (il) [landmark, minimum height=2cm, fill=orange!50] {};

  \draw (i1.north east)+(1.3,0.5) node (conv) [landmark, minimum height=3.2cm, minimum width=1cm, fill=darkgreen!30] {};
  \path (conv.north east)++(1,0.5) node (pool1) [landmark, minimum height=1.5cm, fill=darkgreen!80] {}
    ++(0, -2) node [dot] {}
    ++(0, -0.2) node [dot] {}
    ++(0, -0.2) node [dot] {}
    ++(0, -0.5) node (pool2) [landmark, anchor=north, minimum height=1.5cm, fill=darkgreen!50] {};
  \draw (pool1.north east)+(1.3,-0.5) node (fc) [landmark, minimum height=4.8cm, fill=ggray!20] {};
  \draw (fc.east)+(1,0) node (pred) [landmark, anchor=center, minimum height=1.5cm, line width=1.5pt] {$\mathbf{y}$};

  \draw [decoration={brace,mirror},decorate,semithick] (i1.north -| i4.west)+(-0.2,0) --
    node [anchor=east, text width=1.7cm, align=center] {Landmark features} ($(i4.south west)+(-0.2,0.1)$);
  \draw [decoration={brace,mirror},decorate,semithick] (il.north west)+(-0.2,-0.1) --
    node [anchor=east, text width=1.7cm, align=center] {Local features} ($(il.south west)+(-0.2,0)$);

  \draw [link] (i1.east)+(0.1,0) -- ($(conv.west |- i1.east)+(-0.1, 0)$);
  \draw [link] (i1.east -| conv.east)+(0.1, 0) -- ($(pool1.west)+(-0.1, 0)$);
  \draw [link] (i1.south east -| conv.east)+(0.1, 0) -- ($(pool2.west)+(-0.1, 0)$);
  \draw [link] (pool1.east)+(0.1,0) -- (fc.west |- pool1.east);
  \draw [link] (pool2.east)+(0.1,0) -- (fc.west |- pool2.east);
  \draw [link] (il.east)+(0.1,-0.6) -- ($(fc.west |- il.east)+(0,-0.6)$);
  \draw [link] (fc.east)+(0.1,0) -- (pred.west);
  \draw [link, line width=1.5pt] (pred.south) -- ++(0, -2.5) -- ++(-5.9, 0) -- ($(il.south)+(0,-0.1)$);

  \draw (conv.center) node [rotate=90,text width=3cm,align=center] {Non-Overlapping Convolution};
  \draw (pool1.center) node [rotate=0] {$\Omega_{1}$};
  \draw (pool2.center) node [rotate=0] {$\Omega_{\omega}$};
  \draw (fc.center) node [rotate=90] {Fully-Connected};
  \draw (pred.north)+(0,0.5) node [text width=2cm,align=center] {Coarse prediction};
 
  \hypersetup{linkcolor=black}
  \begin{scope}[on background layer]
    \node [draw=black!70,fill=black!10,fit=(conv) (pool1) (pool2),rounded corners=2pt] (back) {};
    \path (back.south west) node [ggray,anchor=south west] {\emph{\autoref{fig:landpooling}}};
  \end{scope}

  \draw (il.north west)+(-1,0) node [circled] {\textbf{1}};
  \draw (back.north west)+(0.1,-0.1) node [circled,anchor=north west] {\textbf{2}};
  \draw (il.north -| fc.west)+(-0.4,0) node [circled] {\textbf{3}};
  \draw (pred.west)+(-0.3,-0.3) node [circled] {\textbf{4}};
  \draw (back.south)+(0, -1.3) node [circled] {\textbf{5}};

\end{tikzpicture}

%% file: figures_landpooling.tex
\definecolor{darkgreen}{RGB}{0,120,0}
\definecolor{ggray}{RGB}{80,80,80}

\tikzstyle{link}=[->,>=stealth,semithick]
\tikzstyle{landmark}=[draw, minimum width=0.9cm, minimum height=1.2cm, fill=white]
\tikzstyle{filter}=[anchor=west,draw, minimum width=3cm, minimum height=0.6cm, fill=white]

\begin{tikzpicture}[x=1cm,y=0.8cm]

  \path (0, 0) node (l1) [landmark, fill=blue!70] {}
    ++(-0.2,-0.2) node (l2) [landmark, fill=blue!50] {}
    ++(-0.2,-0.2) node (l3) [landmark, fill=blue!30] {}
    ++(-0.2,-0.2) node (l4) [landmark,fill=blue!10] {\footnotesize $\mathbf{x}_{i}[\lambda]$};

  \path (l1)++(2.5,0) node (f1) [filter, fill=darkgreen!70] {}
    ++(-0.2,-0.2) node (f2) [filter, fill=darkgreen!50] {}
    ++(-0.2,-0.2) node (f3) [filter, fill=darkgreen!30] {}
    ++(-0.2,-0.2) node (f4) [filter, fill=darkgreen!10] {};

  \draw [ggray,fill=yellow,fill opacity=0.2] (l1.north east) -- (f1.north west) -- (f1.south west) -- (l1.south east);
  \draw [ggray,fill=yellow,fill opacity=0.2] (l2.north east) -- (f2.north west) -- (f2.south west) -- (l2.south east);
  \draw [ggray,fill=yellow,fill opacity=0.2] (l3.north east) -- (f3.north west) -- (f3.south west) -- (l3.south east);
  \draw [ggray,fill=yellow,fill opacity=0.2] (l4.north east) -- (f4.north west) -- (f4.south west) -- (l4.south east);

  \path (l1)++(2.5,0) node [filter, fill=darkgreen!70] {}
    ++(-0.2,-0.2) node [filter, fill=darkgreen!50] {}
    ++(-0.2,-0.2) node [filter, fill=darkgreen!30] {}
    ++(-0.2,-0.2) node [filter, fill=darkgreen!10] {$\mathbf{F}[\lambda]$};

  \draw (f4.south)+(0, -1) node (gap) [filter, fill=darkgreen!40, anchor=north] {$\mathbf{F}$};

  \path [draw,link] (f4.south)++(-1,0) -- ++(0,-1);
  \path [draw,link] (f4.south)++(0,0) to node [left] {$\Omega$} ++(0,-1);
  \path [draw,link] (f4.south)++(1,0) -- ++(0,-1);
  \path [draw,link,<-,line width=0.5mm] (l4.south) |- ++(-1,-0.4) node [anchor=east] {\textbf{In}};
  \path [draw,link,line width=0.5mm] (gap.south) |- ++(1.5,-0.4) node [anchor=west] {\textbf{Out}};

  \path (f4.west)+(-0.5,0.1) node [anchor=east] {$\mathbf{K}$};

  \path [ggray,draw,|-|] (l4.north west)+(-0.2,0) to node [left] {$k$} ($(l4.south west)+(-0.2,0)$);
  \path [ggray,draw,|-|] (f1.north west)+(0,0.2) to node [auto] {$f$} ($(f1.north east)+(0,0.2)$);
  \path [ggray,draw,|-|] (l1.north west)+(-0.1,0.2) to node [left,yshift=2mm] {$\ell$} ($(l4.north west)+(-0.1,0.2)$);
  \path [ggray,draw,|-|] (f1.south east)+(0.1,-0.2) to node [right,yshift=-2mm] {$\ell$} ($(f4.south east)+(0.1,-0.2)$);

  \begin{scope}[on background layer]
    \node [draw=black!20,fill=black!10,fit=(gap) (f1),rounded corners=2pt] (back) {};
  \end{scope}
\end{tikzpicture}

%% file: tuning.tex
\SetCommentSty{textit}
\SetKwComment{Comment}{$\rhd~$}{}

\begin{algorithm}[tb]
  \DontPrintSemicolon{}
  \SetKwInOut{Input}{input}\SetKwInOut{Output}{output}

  \Input{Predictions $\hat{\gamma}$ and coarse predictions $\mathbf{y}$}
  \Output{Tuned predictions $\hat{\gamma}'$}

  \BlankLine{}
  \Comment{Isolate the best coarse prediction}
  $\phi \leftarrow \argmax(\mathbf{y})$\;
  $p \leftarrow$ \{\emph{indices of features with same family as $\phi$}\}\;\label{tuning:relation}

  \BlankLine{}
  \Comment{Compute the relative weight}
  $w \leftarrow \frac{y_{\phi}}{\sum y_i}$\;\label{tuning:weight}
  $s \leftarrow \sum_{j \in p} \hat{\gamma_{j}}$\;

  \BlankLine{}
  \If{$s = 0 \vee s = 1$}{$\hat\gamma' \leftarrow \hat\gamma$\Comment*[r]{Extreme case}}
  \Else{%

  \lForEach(\Comment*[f]{Bonus})
    {$j \in p$}{$\hat{\gamma_{j}}' \leftarrow
    \hat{\gamma_{j}}\frac{w}{s}$}
    \label{tuning:in}

  \BlankLine{}

  \lForEach(\Comment*[f]{Penalty})
    {$j \notin p$}{$\hat{\gamma_{j}}' \leftarrow
    \hat{\gamma_{j}}\frac{1-w}{1-s}$}
    \label{tuning:out}

  }
  \caption{Multi-label score weighting}\label{alg:tuning}
\end{algorithm}

%% file: figures_deployment.tex
\tikzstyle{link}=[draw,line width=1pt]
\tikzstyle{ovh}=[draw,fill,diamond,blue,minimum size=3pt,inner sep=0pt]
\tikzstyle{scw}=[draw,fill,star,green!30!black,minimum size=3pt,inner sep=0pt]
\tikzstyle{vultr}=[draw,fill,isosceles triangle,orange,minimum size=2.7pt,inner sep=0pt,rotate=90,isosceles triangle apex angle=60]
\tikzstyle{azure}=[draw,fill,circle,red!50!black,minimum size=2.7pt,inner sep=0pt]

\begin{tikzpicture}
  \node (map) at (0, 0) {\includegraphics[width=.3\textwidth]{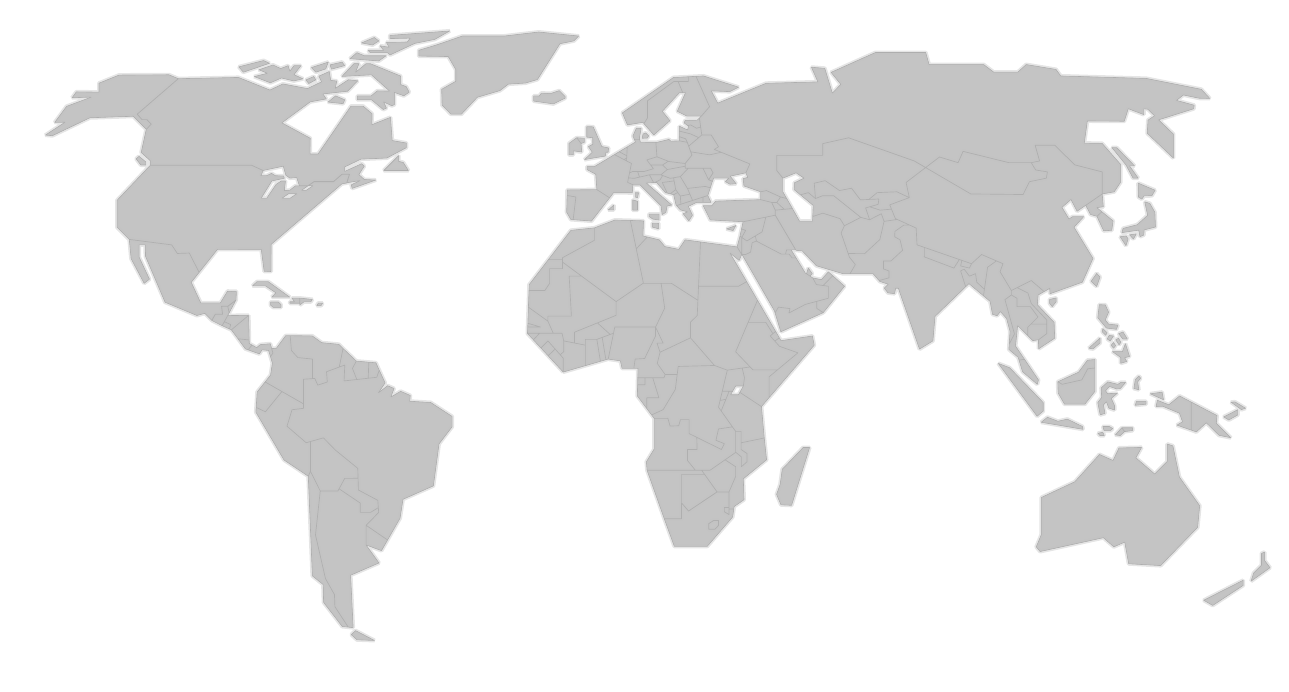}};
  \node (grav) at (-0.17, 0.80) [ovh] {}; 
  \node (pari) at (-0.08, 0.73) [scw] {}; 
  \node (amst) at (-0.05, 0.85) [scw] {}; 

  \node (beau) at (-1.25, 0.78) [ovh] {}; 

  \node (seat) at (-2.1, 0.75) [vultr] {}; 
  \node (west) at (-2.2, 0.64) [azure] {}; 

  \node (miam) at (-1.55, 0.35) [vultr] {}; 
  \node (east) at (-1.50, 0.50) [azure] {}; 

  \node (sing) at (1.60, 0.00) [ovh] {}; 
  \node (asia) at (1.65, -0.09) [azure] {}; 

  \draw (grav)++(-0.07,0.00) node [anchor=east,draw=blue,inner sep=1pt] {\fontsize{6}{6}\selectfont \color{blue} \textsc{grav}};
  \draw (amst)++(0,0.02) node [anchor=west] {\fontsize{6}{6}\selectfont \color{green!30!black} \underline{\textsc{amst}}};
  \draw (pari)++(-0.04,-0.06) node [anchor=west] {\fontsize{6}{6}\selectfont \color{green!30!black} \textsc{pari}};
  \draw (beau)++(0,-0.04) node [anchor=south] {\fontsize{6}{6}\selectfont \color{blue} \underline{\textsc{beau}}};
  \draw (miam)++(0,-0.00) node [anchor=west] {\fontsize{6}{6}\selectfont \color{orange} \textsc{miam}};
  \draw (east)++(0,0.00) node [anchor=west] {\fontsize{6}{6}\selectfont \color{red!50!black} \textsc{east}};
  \draw (seat)++(-0.07,0.03) node [anchor=east,draw=orange,inner sep=1pt] {\fontsize{6}{6}\selectfont \color{orange} \textsc{seat}};
  \draw (west)++(0,-0.03) node [anchor=east] {\fontsize{6}{6}\selectfont \color{red!50!black} \textsc{west2}};
  \draw (sing)++(-0.07,0.00) node [anchor=east,draw=blue,inner sep=1pt] {\fontsize{6}{6}\selectfont \color{blue} \textsc{sing}};
  \draw (asia)++(0.00,0.00) node [anchor=west] {\fontsize{6}{6}\selectfont \color{red!50!black} \textsc{asia}};


  \node (l0) at (-1.9, -1.4) [anchor=east] {\fontsize{6}{6}\selectfont Landmarks:};
  \node (l1) at (-1.8, -1.4) [azure] {};
  \node (l2) at (-0.7, -1.4) [ovh] {};
  \node (l3) at (0.2, -1.4) [scw] {};
  \node (l4) at (1.5, -1.4) [vultr] {};
  \node (ls) at (-0.8, -1.7) [draw=black,inner sep=1pt] {\fontsize{6}{6}\selectfont \textsc{service}};
  \node (ls) at (0.4, -1.7) {\fontsize{6}{6}\selectfont \underline{\textsc{dependency}}};

  \node at (l1.east) [anchor=west] {\fontsize{6}{6}\selectfont Azure};
  \node at (l2.east) [anchor=west] {\fontsize{6}{6}\selectfont OVH};
  \node at (l3.east) [anchor=west] {\fontsize{6}{6}\selectfont Scaleway};
  \node at (l4.east) [anchor=west,yshift=-1.5] {\fontsize{6}{6}\selectfont Vultr};
\end{tikzpicture}

%% file: plots_recalls.tex
\pgfplotsset{%
  discard if not/.style 2 args={%
    x filter/.code={%
      \edef\tempa{\thisrow{#1}}
      \edef\tempb{#2}
      \ifx\tempa\tempb{}
      \else
      \def\pgfmathresult{inf}
      \fi
    }
  }
}%
\begin{tikzpicture}
  \pgfplotsset{%
    small,
    ybar=1pt,
    height=0.35\textwidth,
    width=0.30\textwidth,
    ylabel={Recall@k},
    ymin=0,
    ymax=1.12,
    ytick distance=.2,
    ymajorgrids=true,
    y label style={yshift=-4pt},
    error bars/y explicit,
    error bars/y dir=both,
    error bars/error bar style={gray},
    xtick={0,1,2,3},
    xticklabels={$k=1$,$k=3$,$k=5$,$k=10$},
    xmin=-0.4,
    xmax=3.4,
    x label style={align=center,anchor=north},
    xticklabel style={xshift=2mm,rotate=45,anchor=east},
    nodes near coords,
    nodes near coords align={vertical},
    nodes near coords style={font=\fontsize{6}{7},yshift=2pt,rotate=90,anchor=west,
        /pgf/number format/.cd,fixed,precision=2},
    major grid style={densely dotted},
    legend columns=3,
    legend style={at={(0, 1.02)},anchor=south,font=\scriptsize,draw=none,
    /tikz/every even column/.append style={column sep=1mm}
    },
  }%
  \begin{groupplot}[
    group style={%
      group size=2 by 1,
      yticklabels at=edge left,
      ylabels at=edge left,
      horizontal sep=1mm,
    },
  ]
  \nextgroupplot[bar width=4pt,xlabel={(a) New landmarks}]
    \addplot [blue,pattern=north west lines,pattern color=blue!50!white]
      table [x=i,y=all,y error=all_err,col sep=comma,discard if not={regions}{10}] {data_recalls_gaussian_only_csv.tex};
    \addplot [red,pattern=crosshatch dots,pattern color=red!50!white]
      table [x=i,y=all,y error=all_err,col sep=comma,discard if not={regions}{10}] {data_recalls_forest_only_csv.tex};
    \addplot [darkgreen,fill=darkgreen!20!white]
      table [x=i,y=all,y error=all_err,col sep=comma,discard if not={regions}{10}] {data_recalls_hybrid3p_only_csv.tex};
    \draw [->,black] (rel axis cs:0.04,0.85) -- node [anchor=west,font=\scriptsize] {Better} (rel axis cs:0.04, 0.95);
  \nextgroupplot[bar width=4pt,xlabel={(b) Known landmarks}]
    \addplot [blue,pattern=north west lines,pattern color=blue!50!white]
      table [x=i,y=all,y error=all_err,col sep=comma,discard if not={regions}{10}] {data_recalls_gaussian_except_with_local_csv.tex};
      \addlegendentry{\textsc{Naive Bayes}}
    \addplot [red,pattern=crosshatch dots,pattern color=red!50!white]
      table [x=i,y=all,y error=all_err,col sep=comma,discard if not={regions}{10}] {data_recalls_forest_except_with_local_csv.tex};
      \addlegendentry{\textsc{Random Forest}}
    \addplot [darkgreen,fill=darkgreen!20!white]
      table [x=i,y=all,y error=all_err,col sep=comma,discard if not={regions}{10}] {data_recalls_hybrid3p_except_with_local_csv.tex};
      \addlegendentry{\textsc{DiagNet (ours)}}
  \end{groupplot}
\end{tikzpicture}

%% file: plots_recalls_details.tex
\begin{tikzpicture}
  \pgfplotsset{%
    ybar=1pt,
    small,
    height=0.25\textwidth,
    width=0.47\textwidth,
    x label style={align=center,anchor=south},
    xtick={1,2,3,4,5,6},
    xticklabels={Download,Latency,Loss,Jitter,Uplink,Load},
    ymin=-0.01,
    ymax=1.01,
    ytick distance=.2,
    ylabel={Recall@3},
    major grid style={densely dotted},
    ymajorgrids=true,
    legend columns=3,
    legend style={at={(0.5, 1.02)},anchor=south,font=\scriptsize,draw=none,
    /tikz/every even column/.append style={column sep=1mm}
    },
    error bars/y explicit,
    error bars/y dir=both,
    error bars/error bar style={gray},
  }

  \begin{groupplot}[
    group style={%
      group size=1 by 1,
      xticklabels at=edge bottom,
      xlabels at=edge bottom,
      vertical sep=2mm,
    },
  ]

  \nextgroupplot[bar width=4pt]
  \addplot [blue,pattern=north west lines,pattern color=blue!50!white]
    table [x=i,y=gaussian,y error=gaussian_err,col sep=comma] {data_family_recall_csv.tex};
   \addlegendentry{\textsc{Naive Bayes}}
  \addplot [red,pattern=crosshatch dots,pattern color=red!50!white]
    table [x=i,y=forest,y error=forest_err,col sep=comma] {data_family_recall_csv.tex};
   \addlegendentry{\textsc{Random Forest}}
  \addplot [darkgreen,fill=darkgreen!20!white]
    table [x=i,y=hybrid3p,y error=hybrid3p_err,col sep=comma] {data_family_recall_csv.tex};
    \addlegendentry{\textsc{DiagNet (ours)}}

  \end{groupplot}
\end{tikzpicture}

\begin{tikzpicture}
  \pgfplotsset{%
    ybar=1pt,
    small,
    height=0.25\textwidth,
    width=0.47\textwidth,
    x label style={align=center,anchor=south},
    xtick={1,2,3,4,5,6},
    xticklabels={%
      \textsc{grav$\star$},
      \textsc{seat$\star$},
      \textsc{sing},
      \textsc{beau},
      \textsc{amst},
      local},
    ymin=-0.01,
    ymax=1.01,
    ytick distance=.2,
    ylabel={Recall@3},
    major grid style={densely dotted},
    ymajorgrids=true,
    legend columns=3,
    legend style={at={(0.5, 1.02)},anchor=south,font=\scriptsize,draw=none,
    /tikz/every even column/.append style={column sep=1mm}
    },
    error bars/y explicit,
    error bars/y dir=both,
    error bars/error bar style={gray},
  }

  \begin{groupplot}[
    group style={%
      group size=1 by 1,
      xticklabels at=edge bottom,
      xlabels at=edge bottom,
      vertical sep=2mm,
    },
  ]

  \nextgroupplot[bar width=4pt]
  \addplot [blue,pattern=north west lines,pattern color=blue!50!white]
    table [x=i,y=gaussian,y error=gaussian_err,col sep=comma] {data_region_recall_csv.tex};
  \addplot [red,pattern=crosshatch dots,pattern color=red!50!white]
    table [x=i,y=forest,y error=forest_err,col sep=comma] {data_region_recall_csv.tex};
  \addplot [darkgreen,fill=darkgreen!20!white]
    table [x=i,y=hybrid3p,y error=hybrid3p_err,col sep=comma] {data_region_recall_csv.tex};

    \draw [->,black] (rel axis cs:0.04,0.75) -- node [anchor=west,font=\scriptsize] {Better} (rel axis cs:0.04, 0.95);
  \end{groupplot}
\end{tikzpicture}

%% file: plots_clients.tex
\pgfplotsset{%
  discard if not/.style 2 args={%
    x filter/.code={%
      \edef\tempa{\thisrow{#1}}
      \edef\tempb{#2}
      \ifx\tempa\tempb{}
      \else
      \def\pgfmathresult{inf}
      \fi
    }
  }
}

\begin{tikzpicture}
  \pgfplotsset{%
    small,
    y label style={align=center,anchor=south},
    height=0.20\textwidth,
    width=0.46\textwidth,
    major grid style={densely dotted},
    xmajorgrids=true,
    ymajorgrids=true,
    xlabel=Number of regions with active clients,
    ylabel=Recall@10,
    xtick distance=1,
    ymin=-0.05,
    ymax=1.05,
    error bars/y explicit,
    error bars/y dir=both,
    error bars/error bar style={line width=.5pt,solid},
    error bars/error mark options={line width=.5pt,mark size=2pt,rotate=90},
    legend columns=3,
    reverse legend,
    legend style={at={(0.45, 1)},anchor=south,font=\scriptsize,draw=none,fill=none,
    /tikz/every even column/.append style={column sep=0.3cm}}
  }

  \begin{groupplot}[
    group style={%
      group size=1 by 2,
      xticklabels at=edge bottom,
      xlabels at=edge bottom,
      vertical sep=2mm,
      }
    ]

    \nextgroupplot[ylabel={\footnotesize (a) Failure near \\ \footnotesize new landmarks}]
    \addplot+ [darkgreen, mark=none, discard if not={recall}{5}]
    table [x=regions, y=all, y error=all_err, col sep=comma]
    {data_recalls_hybrid3p_only_csv.tex};
    \addlegendentry{\diagnet}

    \addplot+ [red, densely dashdotted, mark=none, discard if not={recall}{5}]
    table [x=regions, y=all, y error=all_err, col sep=comma]
    {data_recalls_forest_only_csv.tex};
    \addlegendentry{\textsc{Random Forest}}

    \addplot+ [blue, densely dashed, mark=none, discard if not={recall}{5}]
    table [x=regions, y=all, y error=all_err, col sep=comma]
    {data_recalls_gaussian_only_csv.tex};
    \addlegendentry{\textsc{Naive Bayes}}
    
    \draw [->,black] (rel axis cs:0.95,0.75) -- node [anchor=east,font=\scriptsize] {Better} (rel axis cs:0.95, 0.95);

    \nextgroupplot[ylabel={\footnotesize (b) Failure near \\ \footnotesize known landmark}]
    \addplot+ [darkgreen, mark=none, discard if not={recall}{5}]
    table [x=regions, y=all, y error=all_err, col sep=comma]
    {data_recalls_hybrid3p_except_with_local_csv.tex};

    \addplot+ [red, densely dashdotted, mark=none, discard if not={recall}{5}]
    table [x=regions, y=all, y error=all_err, col sep=comma]
    {data_recalls_forest_except_with_local_csv.tex};

    \addplot+ [blue, densely dashed, mark=none, discard if not={recall}{5}]
    table [x=regions, y=all, y error=all_err, col sep=comma]
    {data_recalls_gaussian_except_with_local_csv.tex};

  \end{groupplot}
  
  \path (-1.5cm,0cm) node [anchor=south,rotate=90] {\small Recall@5};
\end{tikzpicture}

%% file: plots_history.tex
\begin{tikzpicture}
  \pgfplotsset{%
      enlarge x limits=false,
      small,
      height=0.22\textwidth,
      width=0.45\textwidth,
      major grid style={densely dotted},
      xmajorgrids=true,
      ymajorgrids=true,
      legend columns=1,
      legend style={at={(0.95, 0.95)},anchor=north east,font=\scriptsize,draw=none,fill=none},
      ymin=0,
  }
  \begin{axis}[
      name=ref,
      xmin=0,
      xmax=39,
      ylabel style={align=center},
      ylabel={General \\ model losses},
    ]

    \node (middle1) at (axis cs:0,0.2) {};

    \addplot+ [draw=blue, mark=none]
    table [x=epoch,y=train_loss,col sep=comma]
    {data_history_0_csv.tex};
    \addlegendentry{Training set}

    \addplot+ [draw=red, mark=none, densely dashdotted]
    table [x=epoch,y=val_loss,col sep=comma]
    {data_history_0_csv.tex};
    \addlegendentry{Validation set}
    
    \node [anchor=east] at (axis cs:37,0.22) {\scriptsize \textsc{Region}/Service};

  \end{axis}

  \begin{groupplot}[
      group style={%
        group size=2 by 3,
        xticklabels at=edge bottom,
        yticklabels at=edge left,
        horizontal sep=5mm,
        vertical sep=3mm,
      },
      at=(ref.below south),
      anchor=north west,
      width=0.255\textwidth,
      ymax=0.15,
      ytick distance=0.1,
      xmax=19,
      height=0.16\textwidth,
    ]

    \nextgroupplot[
      yshift=-5mm,
    ]
    \node (middle2) at (axis cs:0,0) {};
    \addplot+ [draw=blue, mark=none]
    table [x=epoch,y=train_loss,col sep=comma]
    {data_history_5_csv.tex};
    \addplot+ [draw=red, mark=none, densely dashdotted]
    table [x=epoch,y=val_loss,col sep=comma]
    {data_history_5_csv.tex};
    \node at (rel axis cs:0.95,0.95) [anchor=north east,font=\scriptsize] {\textsc{grav}/5};

    \nextgroupplot
    \addplot+ [draw=blue, mark=none]
    table [x=epoch,y=train_loss,col sep=comma]
    {data_history_6_csv.tex};
    \addplot+ [draw=red, mark=none, densely dashdotted]
    table [x=epoch,y=val_loss,col sep=comma]
    {data_history_6_csv.tex};
    \node at (rel axis cs:0.95,0.95) [anchor=north east,font=\scriptsize] {\textsc{grav}/6};

    \nextgroupplot
    \addplot+ [draw=blue, mark=none]
    table [x=epoch,y=train_loss,col sep=comma]
    {data_history_7_csv.tex};
    \addplot+ [draw=red, mark=none, densely dashdotted]
    table [x=epoch,y=val_loss,col sep=comma]
    {data_history_7_csv.tex};
    \node at (rel axis cs:0.95,0.95) [anchor=north east,font=\scriptsize] {\textsc{sing}/1};

    \nextgroupplot
    \addplot+ [draw=blue, mark=none]
    table [x=epoch,y=train_loss,col sep=comma]
    {data_history_10_csv.tex};
    \addplot+ [draw=red, mark=none, densely dashdotted]
    table [x=epoch,y=val_loss,col sep=comma]
    {data_history_10_csv.tex};
    \node at (rel axis cs:0.95,0.95) [anchor=north east,font=\scriptsize] {\textsc{sing}/4};

    \nextgroupplot[
      xlabel={Training epoch},
      every axis x label/.append style={at={(ticklabel cs:1.1,0)}},
      ylabel={Transferred models losses},
      every axis y label/.append style={at={(ticklabel cs:1.8,0)}}
    ]
    \addplot+ [draw=blue, mark=none]
    table [x=epoch,y=train_loss,col sep=comma]
    {data_history_13_csv.tex};
    \addplot+ [draw=red, mark=none, densely dashdotted]
    table [x=epoch,y=val_loss,col sep=comma]
    {data_history_13_csv.tex};
    \node at (rel axis cs:0.95,0.95) [anchor=north east,font=\scriptsize] {\textsc{seat}/2};

    \nextgroupplot
    \addplot+ [draw=blue, mark=none]
    table [x=epoch,y=train_loss,col sep=comma]
    {data_history_14_csv.tex};
    \addplot+ [draw=red, mark=none, densely dashdotted]
    table [x=epoch,y=val_loss,col sep=comma]
    {data_history_14_csv.tex};
    \node at (rel axis cs:0.95,0.95) [anchor=north east,font=\scriptsize] {\textsc{seat}/3};
  \end{groupplot}

  \path (middle1)+(-1.6,0) node {(a)};
  \path (middle2)+(-1.6,-0.3) node {(b)};

\end{tikzpicture}

%% file: diagnet.bbl
\begin{thebibliography}{10}
\providecommand{\url}[1]{#1}
\csname url@samestyle\endcsname
\providecommand{\newblock}{\relax}
\providecommand{\bibinfo}[2]{#2}
\providecommand{\BIBentrySTDinterwordspacing}{\spaceskip=0pt\relax}
\providecommand{\BIBentryALTinterwordstretchfactor}{4}
\providecommand{\BIBentryALTinterwordspacing}{\spaceskip=\fontdimen2\font plus
\BIBentryALTinterwordstretchfactor\fontdimen3\font minus
  \fontdimen4\font\relax}
\providecommand{\BIBforeignlanguage}[2]{{%
\expandafter\ifx\csname l@#1\endcsname\relax
\typeout{** WARNING: IEEEtran.bst: No hyphenation pattern has been}%
\typeout{** loaded for the language `#1'. Using the pattern for}%
\typeout{** the default language instead.}%
\else
\language=\csname l@#1\endcsname
\fi
#2}}
\providecommand{\BIBdecl}{\relax}
\BIBdecl

\bibitem{Sundaresan2013}
\BIBentryALTinterwordspacing
S.~Sundaresan, Y.~Grunenberger, N.~Feamster, D.~Papagiannaki, D.~Levin, and
  R.~Teixeira, ``{WTF? Locating Performance Problems in Home Networks},''
  \emph{CoRR}, 2013. [Online]. Available:
  \url{http://hdl.handle.net/1853/46991}
\BIBentrySTDinterwordspacing

\bibitem{Dimopoulos2015}
G.~Dimopoulos, I.~Leontiadis, P.~Barlet-ros, K.~Papagiannaki, and
  P.~Steenkiste, ``{Identifying the Root Cause of Video Streaming Issues on
  Mobile Devices},'' in \emph{CoNEXT}, 2015.

\bibitem{Joumblatt2013}
D.~Z. Joumblatt, J.~Chandrashekar, B.~Kevton, and R.~Teixeira, ``{Predicting
  User Dissatisfaction with Internet Application Performance at End-Hosts},''
  in \emph{INFOCOM}, 2013.

\bibitem{Nam2016}
H.~Nam, K.~H. Kim, and H.~Schulzrinne, ``{QoE matters more than QoS: Why people
  stop watching cat videos},'' in \emph{INFOCOM}, 2016.

\bibitem{AmazonPLT}
\BIBentryALTinterwordspacing
K.~Eaton. (2012) {How One Second Could Cost Amazon \$1.6 Billion In Sales}.
  [Online]. Available:
  \url{https://www.fastcompany.com/1825005/how-one-second-could-cost-amazon-16-billion-sales}
\BIBentrySTDinterwordspacing

\bibitem{Dischinger2010}
M.~Dischinger, M.~Marcon, S.~Guha, K.~P. Gummadi, R.~Mahajan, and S.~Saroiu,
  ``{Glasnost : Enabling End Users to Detect Traffic Differentiation},'' in
  \emph{Design}, 2010.

\bibitem{Kreibich2010}
C.~Kreibich, N.~Weaver, B.~Nechaev, and V.~Paxson, ``{Netalyzr: Illuminating
  The Edge Network},'' in \emph{IMC}, 2010.

\bibitem{Dhawan2012}
M.~Dhawan, J.~Samuel, R.~Teixeira, C.~Kreibich, M.~Allman, N.~Weaver, and
  V.~Paxson, ``{Fathom: a browser-based network measurement platform},'' in
  \emph{IMC}, 2012.

\bibitem{Sundaresan2011}
S.~Sundaresan, R.~Teixeira, G.~Tech, N.~Feamster, A.~Pescap{\`{e}}, and
  S.~Crawford, ``{Broadband Internet Performance : A View From the Gateway},''
  in \emph{SIGCOMM}, 2011.

\bibitem{Jin2010}
Y.~Jin, N.~Duffield, A.~Gerber, P.~Haffner, S.~Sen, and Z.-L. Zhang,
  ``{NEVERMIND, the problem is already fixed: proactively detecting and
  troubleshooting customer DSL problems},'' in \emph{CoNEXT}, 2010.

\bibitem{LeCun1989}
Y.~LeCun, B.~Boser, J.~S. Denker, D.~Henderson, R.~E. Howard, W.~Hubbard, and
  L.~D. Jackel, ``{Backpropagation Applied to Handwritten Zip Code
  Recognition},'' \emph{Neural Computation}, vol.~1, no.~4, 1989.

\bibitem{Lin2013}
\BIBentryALTinterwordspacing
M.~Lin, Q.~Chen, and S.~Yan, ``{Network In Network},'' \emph{CoRR}, 2013.
  [Online]. Available: \url{https://arxiv.org/abs/1312.4400}
\BIBentrySTDinterwordspacing

\bibitem{Simonyan2014}
\BIBentryALTinterwordspacing
K.~Simonyan, A.~Vedaldi, and A.~Zisserman, ``{Deep Inside Convolutional
  Networks: Visualising Image Classification Models and Saliency Maps},''
  \emph{CoRR}, 2014. [Online]. Available: \url{https://arxiv.org/abs/1312.6034}
\BIBentrySTDinterwordspacing

\bibitem{SpeedtestNetwork}
\BIBentryALTinterwordspacing
Ookla. (2019) {Speedtest Servers}. [Online]. Available:
  \url{https://www.speedtest.net/speedtest-servers}
\BIBentrySTDinterwordspacing

\bibitem{DaHora2018}
D.~N. da~Hora, A.~S. Asrese, V.~Christophides, R.~Teixeira, and D.~Rossi,
  ``{Narrowing the Gap Between QoS Metrics and Web QoE Using Above-the-fold
  Metrics},'' in \emph{PAM}, 2018.

\bibitem{Rubenstein2002}
D.~Rubenstein, J.~Kurose, and D.~Towsley, ``{Detecting shared congestion of
  flows via end-to-end measurement},'' \emph{Transactions on Networking},
  vol.~10, no.~3, 2002.

\bibitem{Duffield2006}
N.~Duffield, ``{Network Tomography of Binary Network Performance
  Characteristics},'' \emph{IEEE Transactions on Information Theory}, vol.~52,
  no.~12, dec 2006.

\bibitem{Dhamdhere2007}
A.~Dhamdhere, R.~Teixeira, C.~Dovrolis, and C.~Diot, ``{NetDiagnoser:
  Troubleshooting network unreachabilities using end-to-end probes and routing
  data},'' in \emph{CoNEXT}, 2007.

\bibitem{Zhao2009}
Y.~Zhao, Y.~Chen, and D.~Bindel, ``{Towards Unbiased End-to-End Network
  Diagnosis},'' \emph{Transactions on Networking}, vol.~17, no.~6, 2009.

\bibitem{Arzani2018}
B.~Arzani, S.~Ciraci, H.~H. Liu, J.~Padhye, B.~T. Loo, G.~Outhred, S.~Design,
  and I.~Nsdi, ``{007: Democratically Finding the Cause of Packet Drops},'' in
  \emph{NSDI}, 2018.

\bibitem{Tan2019}
C.~Tan, Z.~Jin, C.~Guo, T.~Zhang, H.~Wu, K.~Deng, D.~Bi, and D.~Xiang,
  ``{NetBouncer : Active Device and Link Failure Localization in Data Center
  Networks},'' in \emph{NSDI}, 2019.

\bibitem{Tammana2018}
P.~Tammana, C.~Nagarajan, P.~Mamillapalli, R.~Kompella, and M.~Lee, ``{Fault
  localization in large-scale network policy deployment},'' in \emph{ICDCS},
  2018.

\bibitem{Ke2018}
Y.~M. Ke, H.~C. Hsiao, and T.~H.~J. Kim, ``{SDNProbe: Lightweight fault
  localization in the error-prone environment},'' in \emph{ICDCS}, 2018.

\bibitem{Padmanabhan2005}
V.~N. Padmanabhan, S.~Ramabhadran, and J.~Padhye, ``{NetProfiler: Profiling
  wide-area networks using peer cooperation},'' in \emph{IPTPS}, 2005.

\bibitem{He2015}
K.~He, X.~Zhang, S.~Ren, and J.~Sun, ``{Spatial Pyramid Pooling in Deep
  Convolutional Networks for Visual Recognition},'' \emph{TPAMI}, vol.~37,
  no.~9, 2015.

\bibitem{Ribeiro2016}
M.~T. Ribeiro, S.~Singh, and C.~Guestrin, ``{"Why Should I Trust You?":
  Explaining the Predictions of Any Classifier},'' in \emph{SIGKDD}, 2016.

\bibitem{Selvaraju2017}
R.~R. Selvaraju, M.~Cogswell, A.~Das, R.~Vedantam, D.~Parikh, and D.~Batra,
  ``{Grad-CAM: Visual Explanations from Deep Networks via Gradient-Based
  Localization},'' in \emph{ICCV}, 2017.

\bibitem{Madigan1996}
D.~Madigan, A.~E. Raftery, C.~T. Volinsky, and J.~Hoeting, ``{Bayesian Model
  Averaging},'' in \emph{AAAI}, 1996.

\bibitem{linuxtc}
\emph{tc-netem(8) Linux User's Manual}, November 2011.

\bibitem{Arzani2016}
B.~Arzani, S.~Ciraci, B.~T. Loo, A.~Schuster, and G.~Outhred, ``{Taking the
  Blame Game out of Data Centers Operations with NetPoirot},'' in
  \emph{SIGCOMM}, 2016.

\bibitem{Rosenblatt1956}
M.~Rosenblatt, ``{Remarks on Some Nonparametric Estimates of a Density
  Function},'' \emph{The Annals of Mathematical Statistics}, vol.~27, 1956.

\bibitem{Mysore2014}
R.~N. Mysore, R.~Mahajan, A.~Vahdat, and G.~Varghese, ``{Gestalt: Fast, Unified
  Fault Localization for Networked Systems},'' \emph{ATC}, 2014.

\bibitem{Yan2010}
H.~Yan, L.~Breslau, Z.~Ge, D.~Massey, D.~Pei, and J.~Yates, ``{G-RCA: A generic
  root cause analysis platform for service quality management in large IP
  networks},'' in \emph{CoNEXT}, 2010.

\bibitem{Roy2017}
A.~Roy, H.~Zeng, J.~Bagga, and A.~C. Snoeren, ``{Passive Realtime Datacenter
  Fault Detection and Localization},'' \emph{NSDI}, 2017.

\bibitem{Chen2004b}
M.~Chen, A.~Accardi, E.~Kiciman, J.~Lloyd, D.~Patterson, A.~Fox, and E.~Brewer,
  ``{Path-based failure and evolution management},'' in \emph{NSDI}, 2004.

\bibitem{Fonseca2007}
R.~Fonseca, G.~Porter, R.~H. Katz, S.~Shenker, and I.~Stoica, ``{X-trace: A
  pervasive network tracing framework},'' in \emph{NSDI}, 2007.

\bibitem{Giotsas2017}
V.~Giotsas, C.~Dietzel, G.~Smaragdakis, A.~Feldmann, A.~Berger, and E.~Aben,
  ``{Detecting Peering Infrastructure Outages in the Wild},'' in
  \emph{SIGCOMM}, 2017.

\bibitem{Guillot2019}
A.~Guillot, R.~Fontugne, P.~Winter, P.~Merindol, A.~King, A.~Dainotti, and
  C.~Pelsser, ``{Chocolatine: Outage Detection for Internet Background
  Radiation},'' in \emph{TMA}, 2019.

\bibitem{Fok2013}
W.~W.~T. Fok, X.~Luo, R.~Mok, W.~Li, Y.~Liu, E.~W.~W. Chan, and R.~K.~C. Chang,
  ``{MonoScope: Automating network faults diagnosis based on active
  measurements},'' in \emph{IM}, 2013.

\bibitem{Rish2004}
I.~Rish, M.~Brodie, N.~Odintsova, and G.~Grabarnik, ``{Real-time problem
  determination in distributed systems using active probing},'' in \emph{NOMS},
  2004.

\bibitem{Castro2004}
R.~Castro, M.~Coates, G.~Liang, R.~Nowak, and B.~Yu, ``{Network Tomography:
  Recent Developments},'' \emph{Statistical Science}, vol.~19, no.~3, 2004.

\bibitem{Ghita2010}
D.~Ghita, H.~Nguyen, M.~Kurant, K.~Argyraki, and P.~Thiran, ``{Netscope:
  Practical network loss tomography},'' in \emph{INFOCOM}, 2010.

\bibitem{Zhang2018}
Q.~Zhang, C.~Guo, Y.~Dang, N.~Swanson, X.~Yang, R.~Yao, M.~Chintalapati,
  A.~Krishnamurthy, T.~Anderson, and G.~Yu, ``{Deepview: Virtual Disk Failure
  Diagnosis and Pattern Detection for Azure},'' in \emph{NSDI}, 2018.

\bibitem{Steinder2002}
M.~Steinder and A.~Sethi, ``{End-to-end service failure diagnosis using belief
  networks},'' in \emph{NOMS}, 2002.

\bibitem{Bahl2007}
P.~Bahl, R.~Chandra, A.~Greenberg, S.~Kandula, D.~A. Maltz, and M.~Zhang,
  ``{Towards highly reliable enterprise network services via inference of
  multi-level dependencies},'' in \emph{SIGCOMM}, 2007.

\bibitem{Kiciman2008}
E.~Kiciman, D.~Maltz, and J.~C. Platt, ``{Fast Variational Inference for
  Large-scale Internet Diagnosis},'' in \emph{NIPS}, 2008.

\bibitem{Coates2002}
M.~Coates, R.~Castro, R.~Nowak, M.~Gadhiok, R.~King, and Y.~Tsang, ``{Maximum
  likelihood network topology identification from edge-based unicast
  measurements},'' in \emph{SIGMETRICS}, vol.~30, no.~1, 2002.

\bibitem{Huang2008}
Y.~Huang, N.~Feamster, and R.~Teixeira, ``{Practical issues with using network
  tomography for fault diagnosis},'' \emph{SIGCOMM CCR}, vol.~38, no.~5, 2008.

\bibitem{Clark2003}
D.~D. Clark, C.~Partridge, J.~C. Ramming, and J.~T. Wroclawski, ``{A Knowledge
  Plane for the Internet},'' in \emph{SIGCOMM}, 2003.

\bibitem{Wang2004}
H.~J. Wang, J.~C. Platt, Y.~Chen, R.~Zhang, and Y.-M. Wang, ``{Automatic
  Misconfiguration Troubleshooting with PeerPressure},'' in \emph{OSDI}, 2004.

\bibitem{Choffnes2010}
D.~R. Choffnes, F.~E. Bustamante, and Z.~Ge, ``{Crowdsourcing service-level
  network event monitoring},'' in \emph{SIGCOMM}, 2010.

\bibitem{Kim2014}
K.~H. Kim, H.~Nam, V.~Singh, D.~Song, and H.~Schulzrinne, ``{DYSWIS:
  Crowdsourcing a home network diagnosis},'' in \emph{ICCCN}, 2014.

\end{thebibliography}
